\documentclass[times,10pt,twocolumn]{article}

\usepackage{iccv}
\usepackage{times}
\usepackage{epsfig}
\usepackage{graphicx}
\usepackage{amsmath}
\usepackage{amssymb}
\usepackage{times}
\usepackage{epsfig}
\usepackage{graphicx}
\usepackage{tabularx}
\usepackage{subcaption}
\usepackage{booktabs}
\usepackage{datatool}
\usepackage{siunitx}
\usepackage{xspace}
\usepackage{enumitem}
\usepackage{afterpage}
\newcommand\blfootnote[1]{%
  \begingroup
  \renewcommand\thefootnote{}\footnote{#1}%
  \addtocounter{footnote}{-1}%
  \endgroup
}

\usepackage{empheq}
\newcommand*\widefbox[1]{\fbox{\hspace{1em}#1\hspace{1em}}}

\newcommand{\ImNetC}{{ImageNet-C}\xspace}
\newcommand{\ImNetR}{{ImageNet-R}\xspace}
\newcommand{\ImNetClean}{{ImageNet}\xspace}
\newcommand{\ImNetVid}{{ImageNet-Vid}\xspace}
\newcommand{\ImNetVV}{{ImageNet-V2}\xspace}
\newcommand{\ImNetS}{{Stylized-ImageNet}\xspace}
\newcommand{\ObjectNet}{{ObjectNet}\xspace}
\newcommand{\MetaDataset}{{Meta-Dataset}\xspace}

\newcommand{\B}[1]{\textbf{#1}}
\newcommand{\comment}[1]{}

\usepackage[pagebackref=true,breaklinks=true,colorlinks,bookmarks=false]{hyperref}

\newcommand*\dtlformat[1]{\DTLifnumerical{#1}{\num{#1}}{#1}}
\newcommand*\dtlboldmax[2]{
  \DTLforeach{#1}{}{
    \def\theMax{0}
    \DTLforeachkeyinrow{\thisValue}{
      \ifthenelse{\dtlcol>#2}{
        \DTLmax{\theMax}{\theMax}{\thisValue}
      }{}
    }
    \DTLforeachkeyinrow{\thisValue}{
      \ifthenelse{\dtlcol>#2}{
        \ifthenelse{\DTLisieq{\thisValue}{\theMax}}{
          \DTLreplaceentryforrow{\dtlkey}{\textbf{\theMax}}
        }{}
      }{}
    }
  }
} 

\DTLloaddb{meta_dataset}{data/meta_dataset.csv}
\dtlboldmax{meta_dataset}{2}
\DTLloaddb{meta_dataset_test}{data/meta_dataset_test.csv}
\DTLloaddb{imagenet}{data/imagenet.csv}
\DTLloaddb{imagenet_iid_combinations}{data/imagenet_iid_combinations.csv}
\DTLloaddb{clean_corrupt_error}{data/clean_corrupt_error.csv}
\DTLloaddb{data_aug_corruptions}{data/data_aug_corruptions.csv}
\DTLloaddb{imagenetc_our_variations}{data/imagenetc_our_variations.csv}

\iccvfinalcopy 

\def\iccvPaperID{****} 

\ificcvfinal\fi

\begin{document}
\begin{filecontents*}{data/meta_dataset.csv}
DATASOURCE,PREPROCESSING,SUR,BB3,BB5,BB7,BA3,BA5,BA7,ERF3,ERF5,ERF7,STRIDE1
Imagenet,$\downarrow\downarrow\downarrow$,43.31,46.04,46.26,45.56,42.49,41.43,40.76,41.92,40.88,40.75,46.40
Omniglot,$\downarrow$,97.28,97.29,97.62,97.17,97.00,96.71,97.21,96.99,96.90,96.94,97.11
Aircraft,$\downarrow\downarrow\downarrow$,90.11,89.60,90.07,89.35,89.74,89.26,87.62,87.18,87.34,86.58,90.88
Birds,$\downarrow\downarrow\downarrow$,70.87,73.98,71.79,70.29,68.93,68.10,68.34,66.10,65.25,65.50,75.41
Textures,$\downarrow\downarrow\downarrow$,66.98,69.05,68.79,66.58,64.54,65.33,65.09,64.13,61.67,58.77,71.17
Quick Draw,$\uparrow\uparrow$,81.66,82.64,82.46,83.24,81.90,81.66,81.82,82.11,81.23,81.03,82.39
Fungi,$\downarrow\downarrow\downarrow$,65.84,68.56,66.57,66.11,64.25,63.08,62.38,62.25,61.83,62.32,70.95
VGG Flower,$\downarrow\downarrow\downarrow$,86.08,88.55,88.46,86.56,84.58,84.71,83.52,84.61,84.21,83.05,88.60
Average accuracy, ,75.27,76.96,76.50,75.61,74.18,73.79,73.34,73.16,72.41,71.87,77.86
\end{filecontents*}

\begin{filecontents*}{data/meta_dataset_test.csv}
DATASOURCE,SUR,SUR-S,AAS3,AAS3-S,AAS5,AAS5-S,AAS7,AAS7-S,GELU,GELU-S,AASUBG3,AASUBG3-S,AASUBG5,AASUBG5-S,AASUBG7,AASUBG7-S,AASG3,AASG3-S,AASG5,AASG5-S,AASG7,AASG7-S
Imagenet,53.77,1.10,57.32,1.13,58.46,1.08,58.67,1.09,56.81,1.11,59.59,1.08,59.37,1.02,59.06,1.02,59.91,1.06,\textbf{60.59},\textbf{1.04},58.03,1.07
Omniglot,95.81,0.36,96.11,0.36,95.87,0.36,\textbf{96.45},\textbf{0.33},96.29,0.32,96.35,0.32,96.39,0.32,96.25,0.34,96.06,0.33,\textbf{96.45},\textbf{0.31},96.38,0.34
Aircraft,87.47,0.49,89.22,0.43,89.54,0.45,89.90,0.46,88.36,0.56,\textbf{90.93},\textbf{0.40},90.21,0.47,89.98,0.48,90.76,0.46,90.34,0.50,90.68,0.45
Birds,72.44,0.98,78.70,0.87,77.86,0.82,77.36,0.90,72.81,0.89,79.52,0.81,\textbf{80.50},\textbf{0.77},79.76,0.73,78.52,0.81,79.96,0.79,79.51,0.80
Textures,68.96,0.77,71.29,0.90,72.08,0.78,72.77,0.92,72.59,0.80,\textbf{75.00},\textbf{0.75},74.36,0.70,74.55,0.78,74.40,0.73,74.98,0.80,72.61,0.77
QuickDraw,81.58,0.60,82.30,0.54,82.29,0.56,82.29,0.59,83.05,0.54,83.81,0.53,83.23,0.56,83.58,0.52,\textbf{84.01},\textbf{0.54},83.18,0.53,82.88,0.57
Fungi,65.67,1.01,69.87,0.93,71.07,0.98,70.48,0.93,67.91,1.03,72.99,0.95,\textbf{73.86},\textbf{0.92},73.57,0.90,72.88,0.92,73.50,0.89,72.98,0.91
VGG Flower,87.61,0.63,89.04,0.55,87.69,0.60,89.25,0.58,87.55,0.58,90.55,0.47,\textbf{91.24},\textbf{0.47},90.31,0.50,88.81,0.52,89.31,0.48,90.09,0.50
Traffic Signs,51.75,1.06,51.05,1.09,\textbf{55.52},\textbf{1.03},52.15,1.07,53.51,1.06,54.17,1.00,51.00,1.02,51.96,0.99,51.55,1.01,53.51,1.04,49.11,1.00
MSCOCO,48.95,1.10,48.30,1.09,49.61,1.05,47.34,1.09,49.71,1.05,49.53,1.05,49.55,1.01,49.38,1.02,50.28,1.04,\textbf{50.72},\textbf{1.02},47.31,1.02
MNIST,94.04,0.49,92.90,0.55,92.11,0.56,88.06,0.51,\textbf{95.42},\textbf{0.43},94.13,0.46,89.77,0.50,94.05,0.47,93.84,0.51,91.02,0.45,92.46,0.44
CIFAR10,62.45,0.91,66.22,0.84,66.06,0.85,62.32,0.93,63.20,1.01,68.45,0.74,68.21,0.78,66.44,0.77,\textbf{69.60},\textbf{0.77},69.29,0.71,63.75,0.80
CIFAR100,53.12,1.10,56.80,1.04,57.03,1.05,53.63,1.09,56.81,1.21,57.67,1.03,58.50,1.06,58.22,1.02,\textbf{61.58},\textbf{1.04},59.61,1.02,53.40,1.04
Average,71.05,-,73.02,-,73.48,-,72.36,-,72.62,-,\textbf{74.82},-,74.32,-,74.39,-,74.79,-,\textbf{74.80},-,73.01,-
Average {\tiny(in-domain)},76.66,-,79.23,-,79.37,-,79.65,-,78.17,-,81.09,-,\textbf{81.15},-,80.88,-,80.67,-,81.04,-,80.40,-
Average {\tiny(out-of-domain)},62.06,-,63.07,-,64.07,-,60.70,-,63.73,-,\textbf{64.79},-,63.41,-,64.01,-,65.37,-,\textbf{64.83},-,61.21,-
\end{filecontents*}

\begin{filecontents*}{data/imagenet.csv}
WHERE,BB3,BB5,BB7,BA3,BA5,BA7,BBOTH3,BBOTH5,BBOTH7,ERF3,ERF5,ERF7
all layers,       74.20,71.06,68.74,73.91,70.71,68.00,70.92,65.80,61.52,73.06, 71.09,68.67
conv1,     76.18,75.81,75.45,76.10,75.72,75.47,75.76,75.07,74.63,75.27, 75.12,75.03
max pool,  76.00,75.75,75.40,{76.56},76.43,76.35,76.03,75.54,75.24,74.88, 74.87,74.60
block-conv,{76.87},76.74,76.83,{76.88},76.83,76.74,{76.87},76.73,76.63,76.64, 76.71,76.61
strided-skip,      77.05,77.14,77.07,\bf{77.15},77.12,76.83,77.02,76.90,76.62,-,-,-
\end{filecontents*}

\begin{filecontents*}{data/imagenet_iid_combinations.csv}
Method,BB3,BB5,BB7,BA3,BA5,BA7
Subsampled, , , , , , 
Subsampled (low capacity), , , , , , 
\end{filecontents*}

\begin{filecontents*}{data/clean_corrupt_error.csv}
Method,SKIP,MAX-POOL,BLOCK-CONV,CLEAN,GAUSS,SHOT,IMPULSE,DEFOCUS,GLASS,MOTION,ZOOM,SNOW,FROST,FOG,BRIGHT,CONTRAST,ELASTIC,PIXEL,JPEG,MCE
ResNet-50 \tiny{Published} \cite{hendrycks2019robustness}, , , ,23.9,80,82,83,75,89,78,80,78,75,66,57,71,85,77,77,76.7
BlurPool \cite{zhang2019shiftinvar},\checkmark,\checkmark,\checkmark,23.0,73,74,76,74,86,78,77,77,72,63,56,68,86,71,71,73.4
Ours,\checkmark, , ,22.9,68,70,70,72,87,78,75,73,69,50,53,66,81,84,68,70.9
Ours, ,\checkmark, ,23.4,70,72,73,72,87,80,79,72,69,51,54,67,83,84,68,72.0
Ours, , ,\checkmark,23.2,70,72,72,72,88,80,77,73,69,50,53,66,81,84,67,71.6
Ours,\checkmark, ,\checkmark, \textbf{22.5},69,71,72,72,87,79,76,72,68,50,52,64,81,84,68,70.9
Ours,\checkmark,\checkmark, ,22.9,70,71,73,72,80,79,75,71,67,51,53,67,82,85,68,71.2
% anti-alias sub-sampling combined
Ours,\checkmark,\checkmark,\checkmark,\textbf{22.5},68,70,70,72,85,82,75,72,62,50,52,66,81,81,66,\textbf{70.0}
\end{filecontents*}

\begin{filecontents*}{data/data_aug_corruptions.csv}
Method,SKIP,MAX-POOL,BLOCK-CONV,CLEAN,GAUSS,SHOT,IMPULSE,DEFOCUS,GLASS,MOTION,ZOOM,SNOW,FROST,FOG,BRIGHT,CONTRAST,ELASTIC,PIXEL,JPEG,MCE
ResNet-50 \tiny{Published} \cite{hendrycks2019robustness}, , , ,23.9,80,82,83,75,89,78,80,78,75,66,57,71,85,77,77,76.7
BlurPool \cite{zhang2019shiftinvar}, , , ,23.0,73,74,76,74,86,78,77,77,72,63,56,68,86,71,71,73.4
Ours (90 epochs),\checkmark,\checkmark,\checkmark,22.5,68,70,70,72,85,82,75,72,\textbf{62},50,52,66,81,81,66,70.0

AutoAugment**\cite{DBLP:journals/corr/abs-1805-09501}, , , ,22.8,69,68,72,77,83,80,81,79,75,64,56,70,88,57,71,72.7
Rand AutoAugm.**\cite{hendrycks2020augmix}, , , ,23.6,70,71,72,80,86,82,81,81,77,72,61,75,88,73,72,76.1

AUGMIX \cite{hendrycks2020augmix}, , , ,22.4,65,66,67,70,80,66,66,75,72,67,58,58,79,69,69,68.4
ANT (3x3) \cite{rusak2020simple}, , , ,23.9, \textbf{39}, \textbf{40}, \textbf{39}, \textbf{68}, \textbf{78}, 73,77,71,66,68,55, 69,79,\textbf{63},64, \textbf{63.0}
RandAugment** \cite{cubuk2019randaugment}, , , ,22.8,60,58,60,70,90,76,80,70,67,44,50,57,80,86,64,67.4
Ours {\tiny + RandAugm.**},\checkmark,\checkmark,\checkmark,\textbf{21.6}, 59,  58, 61, 70, 84 , 75, 76, 69, 65, \textbf{41}, 48,\textbf{55}, 80, 82,\textbf{61}, 65.5
Swish, , , ,22.4,71,72,74,69,88,80,76,74,69,51,54,68,81,80,67,71.6
Swish + Rand Augm.**, , , ,21.9,61,61,62,69,88,73,78,69,67,42,49,55,81,87,63,66.9
Ours {\tiny+ Swish + Rand Augm.},\checkmark,\checkmark,\checkmark,\textbf{21.2},60,59,61,69,85,\textbf{71},\textbf{75},\textbf{68},\textbf{64},\textbf{41},\textbf{47},\textbf{55},\textbf{78},81,\textbf{61},64.9
\end{filecontents*}

\begin{filecontents*}{data/imagenetc_our_variations.csv}
Method,SKIP,MAX-POOL,BLOCK-CONV,CLEAN,GAUSS,SHOT,IMPULSE,DEFOCUS,GLASS,MOTION,ZOOM,SNOW,FROST,FOG,BRIGHT,CONTRAST,ELASTIC,PIXEL,JPEG,MCE
Baseline, , , ,22.6,70,72,72,71,87,79,76,73,69,51,53,66,81,81,67,71.2
RandAugment** \cite{cubuk2019randaugment}, , , ,22.8,60,58,60,70,90,76,80,70,67,44,50,57,80,86,64,67.4
Ours {\tiny + RandAugm.**},\checkmark, , ,21.8, 60, 60, 60, 70 , 85, 72, 76,69 , 65, 42, 48 , 55 , 80, 86, 62 ,66.2
Ours {\tiny + RandAugm.**},\checkmark,\checkmark,,21.5, 58, 57, 59, 70, 85, 72, 76 , 69, 66, 42, 48 ,56, 78, 82, 62 ,65.2
Ours {\tiny + RandAugm.**},\checkmark, ,\checkmark,21.5, 58, 58, 59, 70, 86, 74, 75,69 , 64, 41,47 ,55 , 79, 83,62,65.4
Ours {\tiny + RandAugm.**},\checkmark,\checkmark,\checkmark,21.6, 59,  58, 61, 70, 84 , 75, 76, 69, 65, 41, 48,55, 80, 82,61, 65.5
Swish, , , ,22.4,71,72,74,69,88,80,76,74,69,51,54,68,81,80,67,71.6
Swish + Rand Augm., , , ,21.9,61,61,62,69,88,73,78,69,67,42,49,55,81,87,63,66.9
Ours {\tiny+ Swish + Rand Augm.},\checkmark, , ,21.4,59,59,59,69,85,74,76,59,65,42,47,55,78,79,62,65.1
Ours {\tiny+ Swish + Rand Augm.},\checkmark,\checkmark, ,21.4,60,59,60,69,85,73,76,67,65,42,47,55,78,82,62,65.3
Ours {\tiny+ Swish + Rand Augm.},\checkmark, ,\checkmark,21.1,60,60,63,69,86,71,75,68,64,41,47,55,78,83,62,65.3
Ours {\tiny+ Swish + Rand Augm.},\checkmark,\checkmark,\checkmark,21.2,60,59,61,69,85,71,75,68,64,41,47,55,78,81,61,64.9
\end{filecontents*}

\title{Impact of Aliasing on Generalization in Deep Convolutional Networks}

\author{
{Cristina Vasconcelos \textsuperscript{1$\dagger$}}
\and Hugo Larochelle \textsuperscript{1,2} 
\and Vincent Dumoulin \textsuperscript{1} 
\and Rob Romijnders \textsuperscript{1} 
\\ \small{\textsuperscript{1} Google Research, Brain Team}
\and  Nicolas Le Roux\textsuperscript{3} 
\\ \small{\textsuperscript{2} Mila, Université de Montréal}
\and Ross Goroshin\textsuperscript{1$\dagger$} 
\\ \small{
    \textsuperscript{3} Mila, McGill University}
}

%

\maketitle

\blfootnote{$\dagger$ \tt\small  \{crisnv,goroshin\}@google.com } 
\begin{abstract} 
We investigate the impact of aliasing on generalization in Deep Convolutional Networks and show that data augmentation schemes alone are unable to prevent it due to structural limitations in widely used architectures. Drawing insights from frequency analysis theory, we take a closer look at ResNet and EfficientNet architectures and review the trade-off between aliasing and information loss in each of their major components. We show how to mitigate aliasing by inserting non-trainable low-pass filters at key locations, particularly where networks lack the capacity to learn them.  
These simple architectural changes lead to substantial improvements in generalization on i.i.d. and even more on out-of-distribution conditions, such as image classification under natural corruptions on ImageNet-C~\cite{hendrycks2019robustness} and few-shot learning on Meta-Dataset~\cite{triantafillou2020metadataset}. State-of-the art results are achieved on both datasets without introducing additional trainable parameters and using the default hyper-parameters of open source codebases. 
\end{abstract}

\section*{Acknowledgement}
\noindent
Hugo Larochelle and Nicolas Le Roux are supported by Canada CIFAR AI Chairs. 

\section{Introduction}

Image analysis in the frequency domain has traditionally played a vital role in computer vision and was even part of the standard pipeline in the early days of deep learning~\cite{Krizhevsky09learningmultiple}. However, with the advent of large datasets, many practitioners concluded that it was unnecessary to hard-code such priors due to the belief that they can be learned from the data itself. \looseness=-1

\begin{figure}[t]
\centering
\begin{minipage}{0.375\linewidth} 
\hfill
\includegraphics[width=0.99\linewidth]{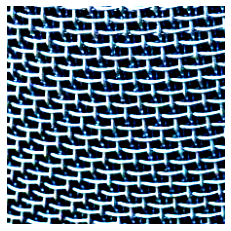}\hfill
\hfill
\end{minipage}
\hfill
\begin{minipage}{0.52\linewidth} 
\includegraphics[width=\linewidth]{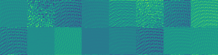} \hfill
\includegraphics[width=\linewidth]{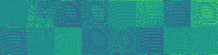} \hfill
\includegraphics[width=\linewidth]{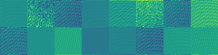}
\end{minipage}
\caption{Input image (Left) and feature maps (Right) from a ResNet-50 illustrating features before subsampling  (top row), aliasing distortions (middle row) and its prevention using anti-aliasing filters (bottom row).
}
\label{fig:aliasing}
\end{figure}

Deep learning approaches thrive in problem settings where labelled data is abundant and training times are virtually unrestricted – allowing the algorithm to apparently learn all necessary features and priors to achieve impressive performance. Central to the success of deep learning approaches on supervised learning problems is the assumption that the training and test data are sampled from the same distribution (i.i.d. conditions). However, many important problem settings involve out-of-distribution (o.o.d.) data or may restrict the amount of labelled data typically required for training deep networks ``from scratch''. In these challenging scenarios, where implicit knowledge cannot be fully obtained from the training datasets, the search for stronger architectural priors as a mechanism to impose explicit knowledge, once again, becomes a critical line of investigation. \looseness=-1

We investigate how aliasing impacts robust generalization under shifts in datasets' spectral distributions and propose simple architectural modifications to mitigate this impact. 
%
%
Our contributions are summarised as follows: (i) A detailed study aimed at more precisely isolating the effects of aliasing in convolutional neural networks (CNNs) is presented. 
(ii) We use frequency analysis theory to derive the critical network locations where anti-aliasing filters must be added, particularly in the pathways that lack the capacity to learn them. We propose simple architectural modifications to ResNet and EfficientNet based models based on these first principles.
(iii) We show that our proposed architecture leads to improved performance on i.i.d. and also on challenging o.o.d. benchmarks, using open source codebases, and their default hyper-parameters.
(iv) We show that this architectural improvement complements other techniques for improving generalization, namely data-augmentation and smooth activations. 
It surpasses the SoTA stand-alone method for \ImNetC in 9 of 15 categories, while producing the lowest clean error on \ImNetClean, and achieves SoTA among approaches that learn on all training sources on \MetaDataset.
 
 


\section{Spectral Aliasing in Convolutional Networks}
\label{sec:questions}

Aliasing is a well known phenomenon that may occur when subsampling any signal.
It occurs when the sampling rate is too low for the signal's bandwidth and does not satisfy the Nyquist rate~\cite{openheim1997signals}. 
Any subsampling that violates this condition causes the high frequency components to additively spill into the signal's low frequency band.
This type of distortion can result in noticeable artifacts .
In classical image/signal processing, aliasing is prevented by applying a low-pass filter before subsampling.
Theoretical fundamentals are presented in \autoref{appendix:DFT}, where Figure \ref{fig:aliasing_theory} illustrates the frequency leakage caused by aliasing.
 
In CNNs, any operation that spatially subsamples its input can potentially cause aliasing, if sampling at a rate lower than twice the highest frequency contained in the input (Nyquist rate).
\autoref{fig:aliasing} illustrates aliasing distortions in CNN's activation maps. It shows the pre-activations and resulting features from a ResNet-50's second subsampling layer, with and without anti-aliasing. 

We now detail the questions addressed in this work.

\begin{itemize}

    \item {\it Are anti-aliasing filters necessary, or do convolutional networks implicitly learn to prevent aliasing?} 
\end{itemize}   
    It is plausible that anti-aliasing filters can be learned by existing, trainable filters of convolutional networks if provided with sufficient spatial support. 
    In Section \ref{sec:methods}, we review the relation between the bandwidth of 
    well known low-pass filters and their spatial support size, 
and establish a relation between CNN filter sizes and their intrinsic capacity to learn low-pass filters directly from data. 
    In Section \ref{sec:arch_review} we use these principles to identify ``aliasing critical paths'' in two modern architectures (ResNet~\cite{DBLP:conf/cvpr/HeZRS16,He2016} and EfficientNet~\cite{DBLP:conf/icml/TanL19})
    and identify their architectural components which are particularly prone to cause aliasing.
    We point to key representation bottlenecks that lack the minimum filter size to represent low-pass filters that prevent these models from learning to prevent aliasing in an end-to-end fashion. We introduce architectural modifications at these bottlenecks which are able to reduce aliasing through low-pass filtering. To prove the contrapositive, we show that introducing these architectural changes where they are not required in theory actually reduces performance in practice. Finally, we experimentally validate the performance improvement of these architectures via ablation studies.  
    
\begin{itemize}    
    \item {\it Can we separate anti-aliasing from other confounding effects?}
\end{itemize}    
    Introducing anti-aliasing filters can affect both prediction and training. Confounding effects with direct influence on performance include: 
    interaction of the anti-aliasing filter with backpropagation dynamics;
    potential increase in receptive field size and 
    indirect smoothing of non-linearities.
    In Section \ref{sec:confounding_effects} we formulate a criterion for optimal placement of low-pass filters and a set of anti-aliasing compositions to isolate the effect of anti-aliasing from these other confounding effects. We perform a number of ablation studies that confirm their differences across a variety of experimental settings.

\begin{itemize}
\item {\it Can standard architectures learn anti-aliasing filters via data-augmentation?} 
\end{itemize}
    Although rich data augmentation can provide an incentive to learn anti-aliasing filters, we show that 
    the performance improvements occur mainly in the low-frequency bands. 
    In contrast, the improvements obtained by our anti-aliased model are shown to benefit features across all spectral bands. We also show that when combined, our model boosts the benefits of data-augmentation to be sustained across the spectrum. Those results confirm our hypotheses that data-augmentation alone cannot prevent anti-aliasing without additional architectural modifications. Finally, we show that combining our model with data augmentation consistently leads to the best results in all of our experiments. In Section \ref{sec:spectral_analysis} we show that this combination extends the invariances induced by data-augmentation across the entire spectrum. 

\begin{itemize}    
    \item {\it Does frequency aliasing impact the generalization performance of convolutional networks?} 
\end{itemize}
    Although input and/or feature map aliasing may occur in deep networks, it is not clear a priori if this phenomenon affects their performance on highly abstract, semantic tasks. 
    Arguably, models can learn to ignore aliased features if in practice they do not correlate with training labels.
    Contradicting this argument, however, we investigate the impact of aliasing on generalization under two hypotheses. First, we claim that a model whose features are susceptible to aliasing may learn brittle correlations that rely on the presence (or absence) of aliasing artifacts in order to generalize. Under this assumption, aliasing impacts out of distribution generalization in settings where test images have different spectral properties, for example, images of varying spatial resolution, compression format, or those affected by natural corruptions. 
    The second, equally important hypothesis, is that aliasing may prevent models from learning useful correlations between features that were corrupted during subsampling. Under this second assumption, we claim that because aliasing can leak frequencies across the entire spectrum, an anti-aliased model has the potential to improve the network's ability to learn useful features that use the entire spectrum. 
    In Section \ref{sec:spectral_analysis} we analyse the performance gains across different spectral bands, when our anti-aliased model is compared to a baseline, and show that it boosts data-augmentation results, as claimed by our second hypothesis. Finally, our ablation studies (Section \ref{sec:results}) evaluate the impact of our proposed model across a large number of datasets (\ImNetClean, \ImNetC, \ImNetR, \ImNetVid, \ImNetVV, \ObjectNet, \ImNetS and another 13 datasets on few-shot-learning) to demonstrate the impact on generalization across a diverse and extremely challenging set of tasks. \looseness=-1


\section{Related Work}
Most relevant to our work is the reformulation of a subsampling layer proposed by Zhang~\cite{zhang2019shiftinvar}. 
They suggest the use of a low-pass filter after a strided-layer's non-linearity, that is a strided-convolution followed by ReLU activation is redefined as non-strided convolution followed by ReLU activation, followed by a strided low-pass filter (\autoref{fig:blur_variations}). 
We point out that this formulation does not fully isolate the aliasing problem from the non-linearity smoothing side-effects, as the filtering operation also smooths the output of the non-linearity. 
%
In contrast, our formulation introduces the filters at the location predicted by aliasing theory. 
Our model delays the removal of high frequencies emerging from non-linear operations until the subsequent subsampling operation. Consequently, our formulation maintains high-frequencies throughout the trainable filters that may exist in between the non-linearity and the next subsampling, but filters them only at the exact point where they may cause aliasing distortion. Zhang's formulation also ignores the network's existing capacity to learn anti-aliasing filters, and because of that, makes use of low-pass filters in layers with sufficient spatial support for learning such filters implicitly end-to-end. 
A direct comparison to Zhang's formulation (\autoref{appendix:ood_generalization}) shows that our model surpasses their results in both i.i.d. and o.o.d. conditions using fewer, and smaller low pass-filters.  
 
Zou et al.~\cite{zou2020delving} propose the use of trainable low-pass filtering layers that operate on feature channel groups and adapt to spatial locations. Their anti-aliasing module augments the model with new trainable and non-linear components, increasing the capacity of the network, thus making it unsuitable for isolating the effects of aliasing from other confounding effects, by the addition of extra convolutional, batch norm, and softmax layers.

Azulay and Weiss~\cite{DBLP:journals/jmlr/AzulayW19} show that CNNs are not as robust to small image transformations as commonly assumed. 
The paper points out that CNN-based models typically ignore the sampling theorem and show large changes in a prediction under small, mostly imperceptible, perturbations of the input. They observe that the improvement in generalization obtained by data augmentation is limited to images that are similar to those seen during training. 
%
Complementary to their findings, we measure the impact of data augmentation in different 
spectral bands and show that improvements in robustness are concentrated in the low frequencies, but at the cost of reducing the model's robustness to changes in mid and high frequency bands. We also show that our model is able to boost the gain associated with data-augmentation to extend its beneficial effects across the entire spectrum. 

Sophisticated augmentation strategies are currently the state-of-the-art approach to o.o.d. classification under natural corruptions~\cite{DBLP:journals/corr/abs-1805-09501,hendrycks2019robustness,hendrycks2020augmix,rusak2020simple}.
Rusak et al.~\cite{rusak2020simple} currently lead the ``standalone leaderboard'' on \ImNetC with a two pass approach. First, they train a generative model to produce additive noise. Next, the classifier and the generative network are jointly trained in an adversarial fashion. Recently, \cite{NEURIPS2019_b05b57f6} pointed out that their improvements are mainly on corruptions that affect high frequencies, while reducing robustness to corruptions that affect low frequencies and also in detriment of ``clean'' (uncorrupted) test accuracy.

While \cite{NEURIPS2019_b05b57f6} pointed out that robustness gains are typically non-uniform across corruption types and that increasing performance in the presence of random noise is often met with reduced performance in corruptions concentrated in different bands, our method overcomes this trade-off (\autoref{sec:imagenetc}). \emph{Note that severe aliasing corrupts the entire spectrum, and not only high frequencies.}

Our results, obtained by combining architectural modifications with off-the-shelf data augmentation (Ekin et al.~\cite{cubuk2019randaugment}) show improvements in all of the 15 corruption categories and outperforms \cite{rusak2020simple} in 9 of them (including fog and contrast, that are concentrated in low frequencies according to \cite{NEURIPS2019_b05b57f6}) while obtaining the highest clean accuracy of the benchmark: $78.8\%$ for \ImNetClean versus $76.1\%$ for \cite{rusak2020simple}. At time of writing, we are not aware of better performing standalone methods on \ImNetC using a ResNet-50 trained on 224$\times$224  \ImNetClean examples only.

\emph{In contrast to the above methods that settle for a trade-off between clean and corrupted accuracy, our method improves both i.i.d. and o.o.d. accuracy simultaneously.}

\section{Methods}
\label{sec:methods}

This section briefly reviews relevant frequency analysis theory, and outlines its implications on CNN architectures. Additional details can be found in Appendix \ref{appendix:DFT}. 

An ideal low-pass filter completely eliminates all frequencies above the cutoff frequency, while allowing those below it to pass without attenuation. It is represented as a rectangular function in the frequency domain and corresponds to the ``sinc'' function in the discrete spatial-domain with support size equal to the number of elements on the input itself. The spatial support size is defined by the interval with minimum length containing all nonzero filter weights.
Learning a good approximation of the ideal low-pass filter requires much larger filters than those typically found in the individual layers of modern CNNs. 

A non-ideal low-pass filter has transition regions, defined as intervals between its passband and stopband responses in the frequency domain. Although they may have much smaller spatial support size, the existence of transition regions imposes a trade-off between preserving information and reducing aliasing.

Any convolutional layer that represents a low-pass filter must have a filter of size of at least 2. A discrete filter with size 1 can only represent an impulse function. Its Fourier transform is constant across the spectrum, which scales all frequencies equally, thus it cannot act as a low pass filter. Convolutional layers with filter size larger than 1 are able to represent low-pass filters, up to varying degrees of approximation error for different bandwidths. However, the uncertainty principle implies that a signal's spatial support size is inversely proportional to its spectral support size \cite{parhizkar2015sequences}. For this reason, the larger the filter size of a convolutional filter, the better it can approximate a narrow low-pass filter.



Another important property pertains to a stack of convolutional layers which may learn to prevent aliasing in a distributed fashion. More formally, a filter is said to be separable if it can be expressed as the composition of two or more filters. A stack of convolutional layers with small filter sizes (but larger than 1) may learn to represent a larger, separable, low-pass filter.

\subsection{Optimal Placement and Confounding effects}
\label{sec:confounding_effects}

The design principle of our anti-aliasing models trade-off between two opposing requirements: band-limiting the spectrum of subsampled feature maps in order to prevent aliasing in layers that subsample, while simultaneously preserving the information encoded in high frequencies as much as possible throughout the other layers.

One naive extreme of this balance is to pre-process the input image with a low-pass filter that has very low cut-off frequency. This solution is sub-optimal for two reasons: (i) it induces early loss of information that may be crucial for solving the task (e.g. classification); (ii) non-linearities may introduce high-frequencies in the internal feature maps, causing them to be susceptible to aliasing in subsequent subsampling layers. 

We claim that the solution proposed in \cite{zhang2019shiftinvar} is sub-optimal.
It suggests the use of a low-pass filter just after a subsampling layer non-linearity. We observe that the high frequencies created by this non-linearity are harmless up to the next subsampling operator, and thus this solution prematurely discards high frequency information.
To satisfy the two requirements mentioned above, an optimal anti-aliased model should be able to preserve high-frequencies up to the subsequent subsampling operation. To allow the network's features to make use of high frequency information while avoiding aliasing, \emph{it is ideal to apply low-pass filtering immediately before subsampling}.   

\begin{figure}[th]
\centering
\includegraphics[width=\linewidth]{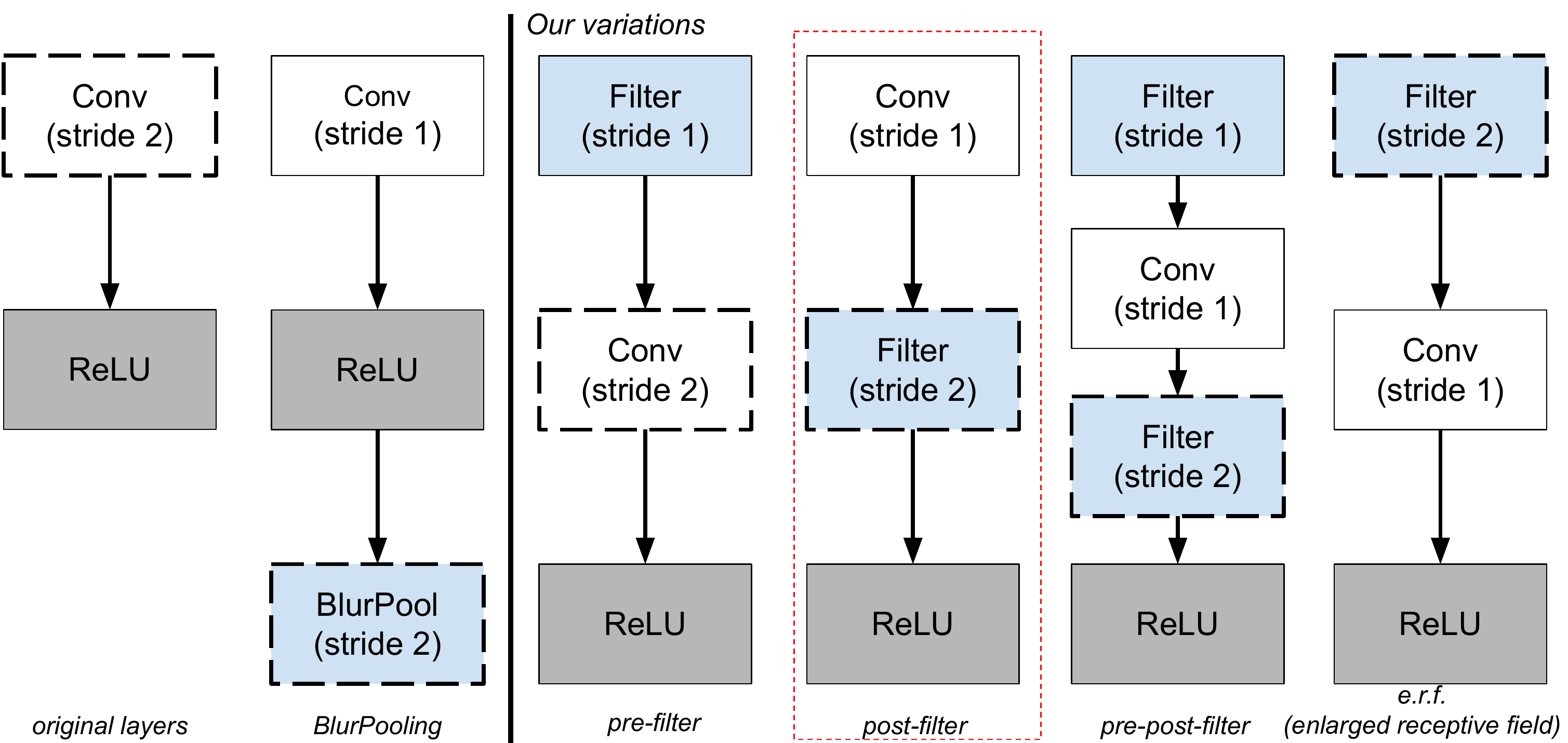}
\caption{\label{fig:blur_variations}
Low-pass filters can be added at various locations with potentially different side-effects. {\em BlurPooling}~\cite{zhang2019shiftinvar} suggest their addition after the ReLU non-linearity, whereas our variants apply low-pass filtering before the original subsampling and preserve any high frequencies produced by the non-linearity. The optimal ``post-filter'' variation is outlined in red. 
}
\end{figure}

Optimizing these criteria does not uniquely specify an anti-aliasing architecture. All four variations in Figure \ref{fig:blur_variations} satisfy the stated optimality criteria. Nevertheless, each variation introduces its own side-effects on training, both in the forward and backward pass. 
Although convolutions \emph{per se} are commutative (see \cite{book_097456074X} for a detailed proof), note that this property no longer holds when back-propagating gradients or when the layers include subsampling/non-linearities. That is, a change in the order of the fixed and trainable filters induces different priors. For example, the \emph{backward} pass in the \emph{post-filter} variant induces the trainable convolution filter to be influenced by a larger neighborhood of gradients than the \emph{pre-filter} variant. Similarly, the decision of where to include subsampling affects the \emph{resolution} of the feature maps and also the gradients that influence the trainable layer. In our variant named \emph{enlarged receptive field (e.r.f.)}, we explore the case where subsampling is part of the low-pass filter that precedes the trainable convolution, which potentially increases the receptive field size of the trainable filter as a side effect.

Among the four variations, the {\em post-filter} (outlined in red in  \autoref{fig:blur_variations}) has the benefit of smoothing gradients with the low-pass filter in the backward pass, in a way that better aligns the low-pass filter with the positions where gradients are up-sampled.  
This potentially explains why this variant produces superior empirical results, as confirmed in \autoref{table:imagenet}.


\subsection{Aliasing critical paths}
\label{sec:arch_review}
In this section we characterize a network's capacity to learn anti-aliasing filters end-to-end. 
Although low-pass filtering can occur in any of the trainable convolution layers, in order for low-pass filters to prevent aliasing they must be located between the operation that produces high frequencies and the subsampling operation. For this reason, we define the aliasing critical path as the set of operations preceding subsampling that maintain or reduce the range of frequencies contained in the signal and lack the capacity to produce new high frequency content. The burden of preventing aliasing, in accordance with the sampling theorem, falls on the operations found in the critical path.
Given a sequence of layers, the aliasing critical path for a certain subsampling operation traces back to the last non-linearity that precedes it (or the input image itself if none exist). 
Non-linearities may produce high-frequencies (see appendix \ref{appendix:smooth}), but these are harmless up until the next down-sampling operation.
Skip connections are defined as the summation of a feature map with a downstream feature map. 
They can act as pathways for replicating frequencies through the network, bypassing any low-pass filters that may have been learned in the main path \footnote{In theory, the summation at the skip connection can act as a low-pass filter through phase cancellation. Phase cancellation occurs when two (or more) signals of the same frequency but with inverted phases (a phase difference of $\pi$) cancel each other when summed together.}. 

In summary, we define the critical path for learning low-pass filters as the sequence of operations between a subsampling operation and either the skip connection or a non-linearity preceding it, whichever is closest to the subsampling operation.
Note that our definition of critical path for learning anti-aliasing does not preclude trainable layers further upstream (i.e. preceding the skip-connection or non-linearity) from serving as anti-aliasing filters by aggressively low-pass filtering their input. 
Rather, we define a critical set of layers that must act as ``last chance'' low-pass filters, whenever the incoming signal does not  satisfy the Nyquist rate.



\subsection{ResNets and EfficientNets}

\begin{figure}[t]
\centering
\includegraphics[width=0.98\linewidth]{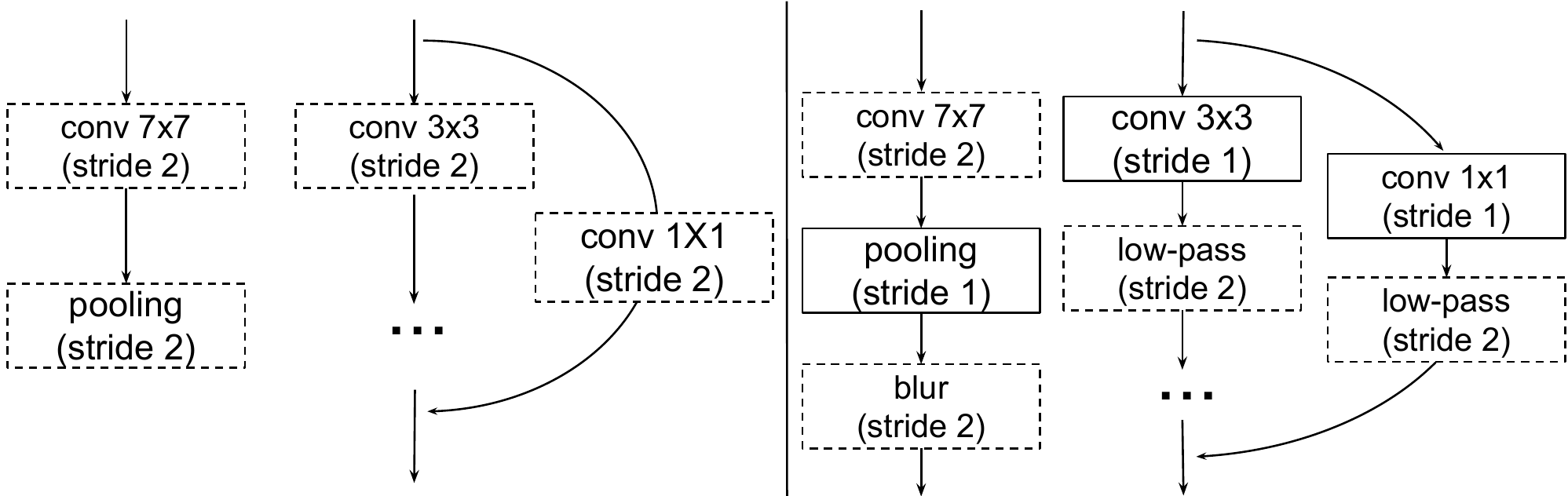}
\caption{
ResNets components that include subsampling. 
Strided-skip connections filters (1 $\times$ 1) lack the minimum size necessary to represent a low pass filter.
Left: Original composition. Right: Our anti-aliased model based on applying the \emph{post-filter} composition only at key places. Non-linearities are omitted for clarity. Dashed lines denote layers with subsampling. 
}
\label{fig:closer_look}
\end{figure}
ResNet's aliasing critical paths can be differentiated according to their composition into the following: (1) the initial strided-convolutional layer; (2) the initial strided-max-pooling (absent in some implementations such as~\cite{dvornik2020selecting}); (3) the sequence of convolutions within a strided-residual block main path up to its subsampling operation; and (4) the sequence of convolutions within a strided-skip-connection up to its spatial subsampling. \autoref{fig:closer_look} illustrates the components in these critical paths and their filter sizes which impact their capacity to prevent aliasing. 
The right-hand side of the figure illustrates our proposed model, respecting the optimal placement criteria (Section \ref{sec:confounding_effects}). Note that the $7 \times 7$ trainable convolution has sufficient capacity to act as an anti-aliasing filter, therefore our model does not include an additional anti-aliasing filter in this critical path.


\begin{figure}[t]
\centering
\includegraphics[width=0.4\linewidth]{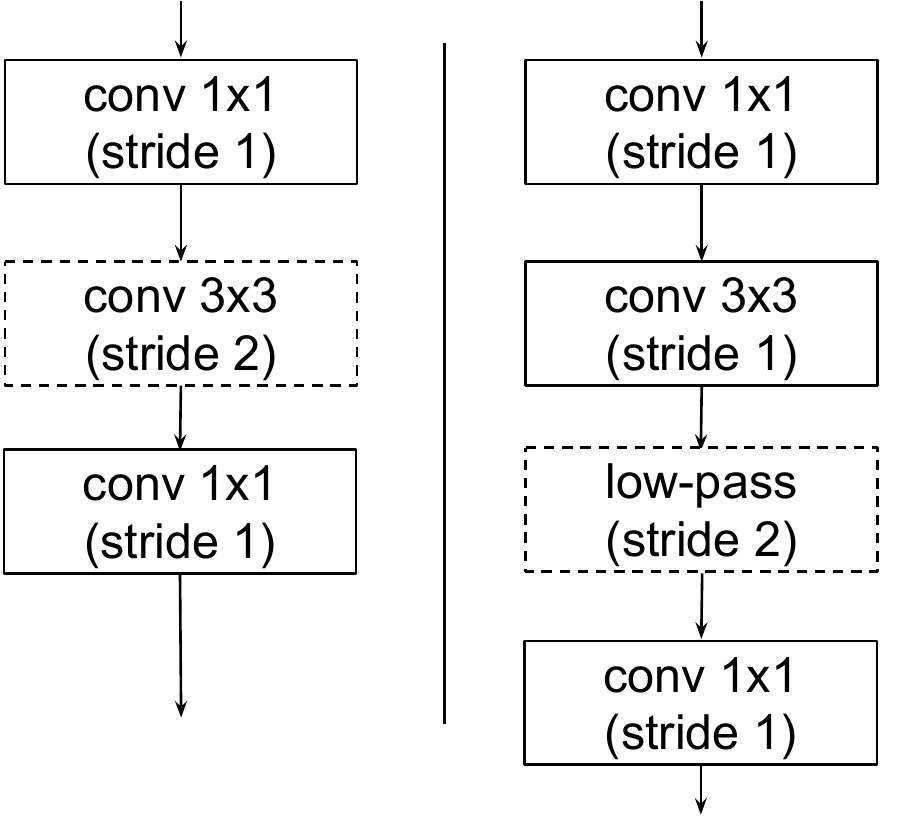}
\caption{EfficientNet layers that include subsampling.  Left: Original composition. Right: Our anti-aliased model which uses the \emph{post-filter} variation at key locations.
Non-linearities are omitted for clarity. Dashed lines indicate layers with subsampling.}
\label{fig:efficientnets}
\end{figure}

EfficientNet is currently the best-performing architecture in several resource constrained settings \cite{DBLP:conf/icml/TanL19}. Its baseline architecture was found via neural architecture search, which optimizes both accuracy and efficiency (FLOPS), while balancing the network's depth, width, and input and feature map resolution as hyper-parameters. It is interesting to analyse the EfficientNet architecture from the perspective of aliasing because the hyper-parameters that were optimized are directly related to the aliasing problem.

%
EfficientNets's aliasing critical paths can be differentiated according to their composition into the following (\autoref{fig:efficientnets}): 
(1) initial strided-convolutional layer; (2) strided-blocks with a sequence of 1x1 and 3x3 convolutions. Layers that perform subsampling do not contain a parallel path with skip connections, while other blocks do.
%
 %
EfficientNets use the Swish activation function ~\cite{ramachandran2017searching} instead of ReLU. 
Smooth functions have rapidly decaying Fourier transforms \cite{katznelson_2004,bresler2020corrective}. 
This property reduces the likelihood, assuming random inputs, of introducing high frequency content as compared to non-smooth activation functions (see  Appendix \ref{appendix:smooth}). 


\section{Experimental Results} 
\label{sec:results}

\begin{table*}[ht]
\scriptsize
\begin{center}
\begin{tabular}{@{}lccccccccccccccc@{}}
  \toprule
  & \multicolumn{3}{c}{Pre-filter {\tiny (filter size $k$)}} & \phantom{a} & \multicolumn{3}{c}{Post-filter {\tiny (filter size $k$)}} & \phantom{a} & \multicolumn{3}{c}{Pre-post-filter {\tiny (filter size $k$)}} & \phantom{a} & \multicolumn{3}{c}{Enlarge receptive field {\tiny (filter size $k$)}} \\ 
  \cmidrule{2-4} \cmidrule{6-8} \cmidrule{10-12} \cmidrule{14-16}
  Location & $k=3$ & $k=5$ & $k=7$ && $k=3$ & $k=5$ & $k=7$ && $k=3$ & $k=5$ & $k=7$ && $k=3$ & $k=5$ & $k=7$\\
  \midrule
  \DTLforeach{imagenet}{%
    \where=WHERE,%
    \bbthree=BB3,%
    \bbfive=BB5,%
    \bbseven=BB7,%
    \bathree=BA3,%
    \bafive=BA5,%
    \baseven=BA7,%
    \bboththree=BBOTH3,%
    \bbothfive=BBOTH5,%
    \bbothseven=BBOTH7,%
    \erfthree=ERF3,%
    \erffive=ERF5,%
    \erfseven=ERF7%
  }{%
    \DTLiffirstrow{}{\\}%
    \dtlformat{\where}&%
    \dtlformat{\bbthree}&%
    \dtlformat{\bbfive}&%
    \dtlformat{\bbseven}&&%
    \dtlformat{\bathree}&%
    \dtlformat{\bafive}&%
    \dtlformat{\baseven}&&%
    \dtlformat{\bboththree}&%
    \dtlformat{\bbothfive}&%
    \dtlformat{\bbothseven}&&%
    \dtlformat{\erfthree}&%
    \dtlformat{\erffive}&%
    \dtlformat{\erfseven}
  }
  \\ \bottomrule
\end{tabular}



\end{center}
\caption{\label{table:imagenet}
Imagenet results: rows demonstrate the impact of anti-aliasing the model's components from \autoref{fig:closer_look} individually 
while columns show blur variations from \autoref{fig:blur_variations}. Results show 
a significant accuracy increase when anti-aliasing the strided-skip connections of a Resnet-50 and the negative impact of information loss when blur is applied to all layers indistinctly and also to the first convolutional layer that contain  large  trainable  kernels, capable  of learning anti-aliasing filters.
Note that baseline accuracy is $76.49\%$. The values correspond to the mean accuracy over 3 runs with different seeds.
}
\end{table*}

Our experiments investigate different aspects of the impact of aliasing on performance. We begin by evaluating the i.i.d. generalization performance of our proposed models on the \ImNetClean benchmark in Sections \ref{sec:imagenet_results}-\ref{sec:spectral_analysis}. 
The experiments in \autoref{sec:imagenet_results} are designed to validate the hypotheses presented in \autoref{sec:methods} such as the optimal placement of low-pass filters. These experiments are also designed to disambiguate the improvements obtained by anti-aliasing from other confounding effects.

Next, the experiments in \autoref{sec:spectral_analysis} show that anti-aliasing improves the results obtained by data-augmentation by extending its effects across a larger range of frequencies.

Finally, we evaluate our models on challenging o.o.d. benchmarks and show that our proposed modifications lead to even more striking performance improvement -- achieving state-of-the art performance without any additional hyper-parameter sweeps.
The detailed description of these datasets is presented in the appendices (see \autoref{appendix:imagenetc} for \ImNetC, \autoref{appendix:metadataset} for Meta-Dataset, and \autoref{appendix:ood_generalization} for extra results on other datasets used in robustness analysis).  
Our experiments use TensorFlow's official ResNet-50 and EfficientNet-B0 models\footnote{\scriptsize{\url{https://github.com/tensorflow/models/tree/master/official/vision/image\_classification/resnet}}}.
The o.o.d. experiments on the Meta-Dataset benchmark use the publicly available SUR codebase ~\cite{dvornik2020selecting} -- a recent few-shot classification model. 
On the following tables, results on bold implies ``better'' with statistical significance.


\subsection{ImageNet}
\label{sec:imagenet_results}

To investigate the effects of aliasing in the ResNet family of architectures we initially evaluate their performance on \ImNetClean~\cite{ILSVRC15}. Our results show that ResNets are most severely impacted by aliasing in their strided skip connections (which are preceded by $1 \times 1$ convolutions). 

We used the official TensorFlow \cite{tensorflow2015-whitepaper} public code for training a ResNet-50 architecture, yielding a top-1 accuracy of $76.49\%$. The codebase reproduces the training pipeline and hyper-parameters from \cite{GoyalDGNWKTJH17}, in which models are trained for 90 epochs. 
\autoref{table:imagenet} shows the effect on top-1 accuracy of adding anti-aliasing filters before (\emph{pre}), after (\emph{post}), and before-and-after (\emph{pre-post}), various operations in the network.
Recall that adding these fixed filters may also affect the back-propagation of gradients, as well as increase the receptive field size. Our ablation studies are designed to disambiguate these effects from anti-aliasing. 
\autoref{table:imagenet} reports accuracies as a result of adding a \emph{pre-filter} (77.14\%) or \emph{post-filter} (77.15\%) to the skip connections \emph{that include subsampling}, and are immediately preceded by a trainable $1 \times 1$ convolutional layer. 
Note that inserting low-pass filters into all convolutional layers (``all'' row) degrades the performance to as low as 61.00\% accuracy. The second row shows that anti-aliasing the ``conv1'' layer degrades performance because it already has the capacity to learn a low-pass filter in its $7 \times 7$ kernel.
The last column, ERF (enlarged receptive field), is aimed to disambiguate whether the improved performance is truly due to anti-aliasing or merely the enlarged receptive field caused as a side-effect. 

Finally, by combining anti-aliasing with \emph{post-filter} variants at all strided layers except the first strided-convolution which has large filter size (\autoref{fig:closer_look}), we are able to report a top-1 accuracy of (77.47\%). These results support our hypothesis that anti-aliasing in networks must trade-off between mitigating the effects of aliasing while preserving high-frequency information as much as possible. This is achieved by adding anti-aliasing filters only to those critical paths that lack the capacity to learn them.

Finally we note that analogous improvements were also obtained with EfficientNets using the same design principles. These results can be found in \autoref{appendix:efficientNets}.

\subsection{Data-Augmentation}
\label{sec:spectral_analysis}

\begin{figure*}[ht]
\centering
\begin{minipage}{0.47\textwidth}
\includegraphics[width=\linewidth]{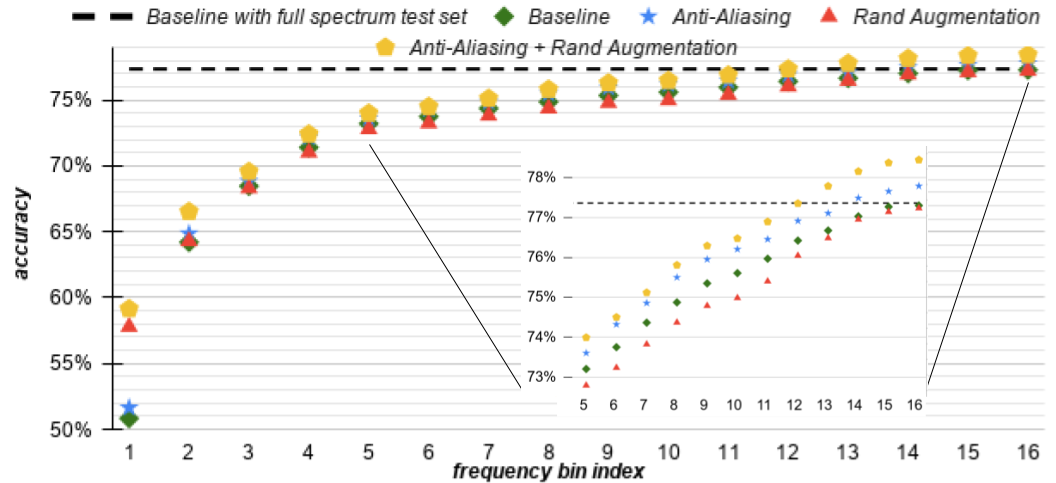}
\end{minipage}
\begin{minipage}{0.47\textwidth}
\includegraphics[width=\linewidth]{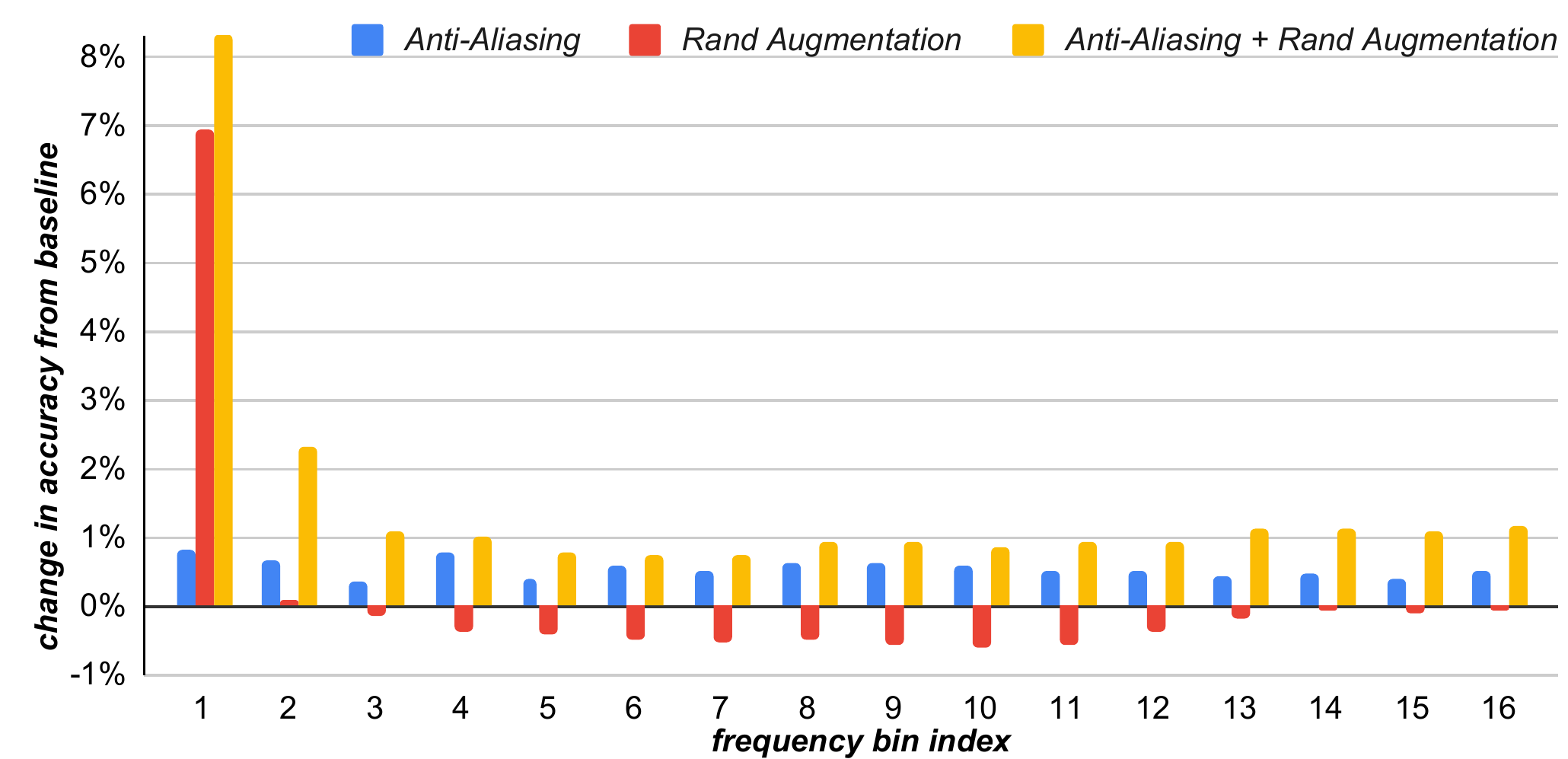}
\end{minipage}
\caption{\label{fig:robustness_to_frequency_removal} The impact on pre-trained models when tested on images that have 1/16 of their spectral band removed.
\emph{Left:} Degradation Curve. All models are more robust to the lack of high-frequency than low frequency content. \emph{Right} Difference between the baseline performance and the remaining three models. 
Our anti-aliased model performance is higher than baseline in all spectral bands, presenting positive differences across the spectrum.
Data augmentation model performance is higher when removing content from the lowest frequencies, but presents the worst performance of all models for the remaining bins.  
The combined model (AA+DA) boosts their individual advantages.}
\end{figure*}

\begin{figure}[ht]
\centering
\captionsetup[subfigure]{labelformat=empty}
\begin{minipage}{0.99\linewidth}
\centering
\includegraphics[width=0.12\textwidth]{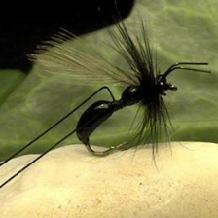}\hfill
\includegraphics[width=0.12\textwidth]{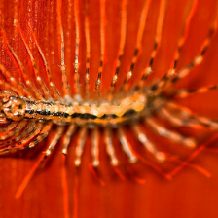}\hfill
\includegraphics[width=0.12\textwidth]{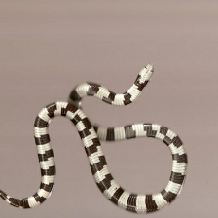}\hfill
\includegraphics[width=0.12\textwidth]{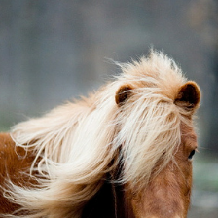}\hfill
\includegraphics[width=0.12\textwidth]{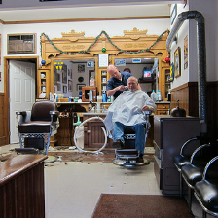}\hfill
\includegraphics[width=0.12\textwidth]{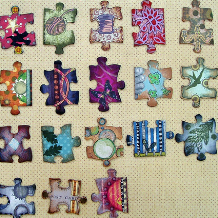}\hfill
\includegraphics[width=0.12\textwidth]{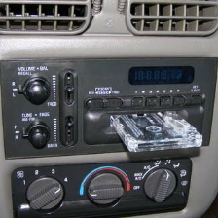}\hfill
\includegraphics[width=0.12\textwidth]{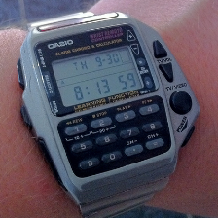}\\
\includegraphics[width=0.12\textwidth]{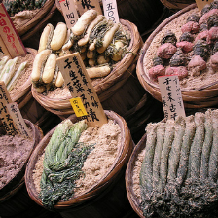}\hfill
\includegraphics[width=0.12\textwidth]{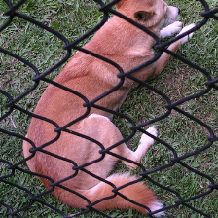}\hfill
\includegraphics[width=0.12\textwidth]{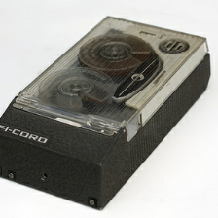}\hfill
\includegraphics[width=0.12\textwidth]{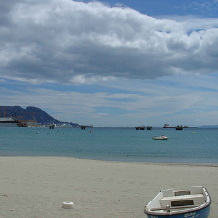}\hfill
\includegraphics[width=0.12\textwidth]{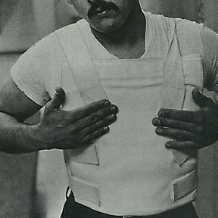}\hfill 
\includegraphics[width=0.12\textwidth]{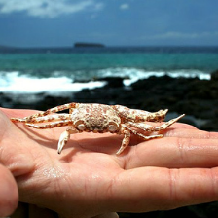}\hfill
\includegraphics[width=0.12\textwidth]{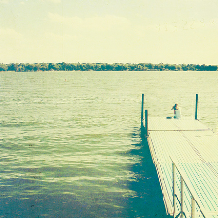}\hfill 
\includegraphics[width=0.12\textwidth]{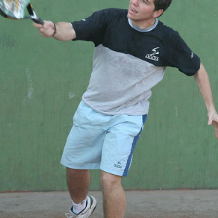}\\
\includegraphics[width=0.12\textwidth]{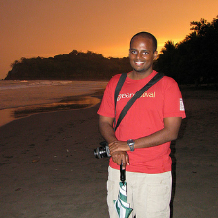}\hfill
\includegraphics[width=0.12\textwidth]{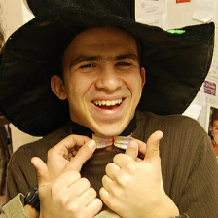}\hfill
\includegraphics[width=0.12\textwidth]{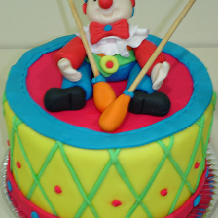}\hfill
\includegraphics[width=0.12\textwidth]{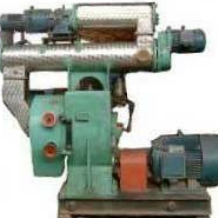}\hfill
\includegraphics[width=0.12\textwidth]{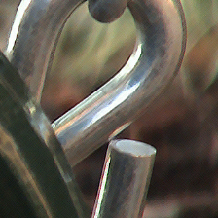}\hfill
\includegraphics[width=0.12\textwidth]{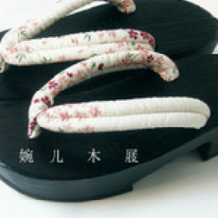}\hfill  
\includegraphics[width=0.12\textwidth]{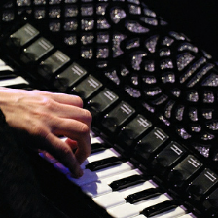}\hfill
\includegraphics[width=0.12\textwidth]{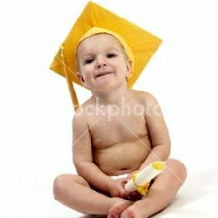}\\
\includegraphics[width=0.12\textwidth]{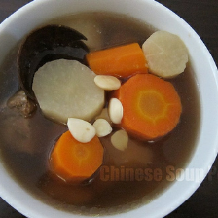}\hfill
\includegraphics[width=0.12\textwidth]{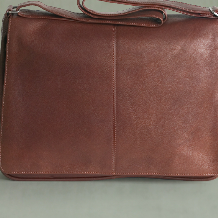}\hfill
\includegraphics[width=0.12\textwidth]{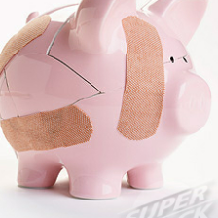}\hfill
\includegraphics[width=0.12\textwidth]{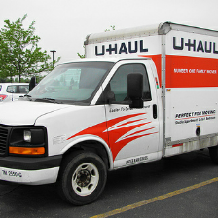}\hfill
\includegraphics[width=0.12\textwidth]{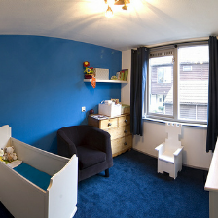}\hfill
\includegraphics[width=0.12\textwidth]{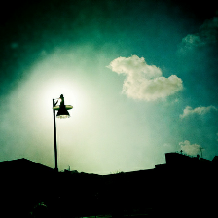}\hfill
\includegraphics[width=0.12\textwidth]{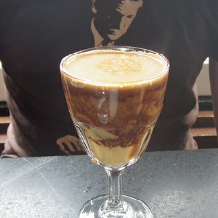}\hfill
\includegraphics[width=0.12\textwidth]{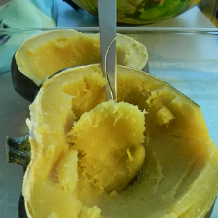}
\comment{
\includegraphics[width=0.12\textwidth]{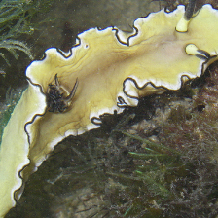}\hfill
\includegraphics[width=0.12\textwidth]{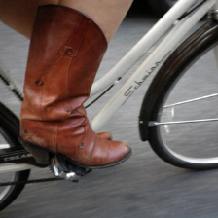}\hfill
\includegraphics[width=0.12\textwidth]{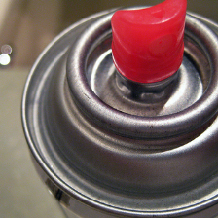}\hfill
\includegraphics[width=0.12\textwidth]{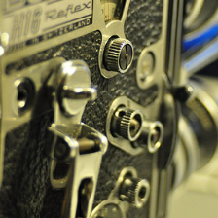}\hfill
\includegraphics[width=0.12\textwidth]{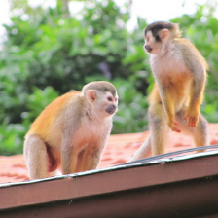}\hfill
\includegraphics[width=0.12\textwidth]{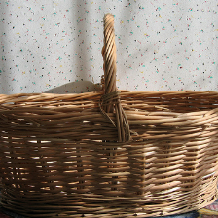}\hfill
\includegraphics[width=0.12\textwidth]{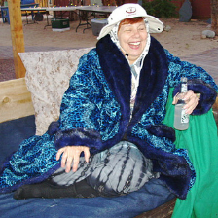}
}
\end{minipage}
\caption{\label{fig:images_frequency_removal1} Random subset of \ImNetClean wrongly classified by both baseline and data-augmented models, but correctly classified by our Anti-Aliased model.
The images have different spectral characteristics, from those that include high frequencies (fine structures and texture) to those with low frequency content (regions of constant color).}
\end{figure}

The ablation studies presented in this section are designed to differentiate between performance gains obtained by data-augmentation from those obtained from our anti-aliased model.
The four models evaluated in this section and their respective accuracies on \ImNetClean are: (i) ``Baseline''-baseline model (77.36\%), (ii) ``Anti-Aliasing''-our proposed model containing non-trainable anti-aliasing filters (77.76\%), (iii) ``Rand Augmentation''- model trained with  random data augmentation \cite{cubuk2019randaugment} (77.32\%), (iv) ``Rand Augmentation + Anti-Aliasing'' combines (ii) and (iii) (78.45\%).
These models were trained with the same number of epochs (180) on \ImNetClean only, as the public source code from \cite{cubuk2019randaugment} (when data-augmentation is enabled).

Figure \ref{fig:images_frequency_removal1} illustrates images correctly classified by our anti-aliased model, but not by the baseline nor the model trained with data-augmentation.
Note that these images contain different spectral characteristics suggesting that anti-aliased models benefit features across the entire spectrum.
To further explore this observation we investigate the reliance of trained networks on various frequency bands of input images. This can be done by measuring the impact of removing frequency intervals from the input images at inference time only.
To accomplish this, we divide the input spectrum into 16 bands and apply a ``notch filter'', zeroing out one frequency interval at a time.
We then evaluate the performance of the same four models (trained on unfiltered images) on the resulting filtered image test sets. The results are presented in Figure \ref{fig:robustness_to_frequency_removal}. 
The left plot shows that for all models, the performance impact is larger when lower frequency bands are filtered out. The right plot shows the change in accuracy from the baseline accuracy over the same bin. Notice a striking characteristic pattern of decreased performance around the mid-band frequencies of the ``Rand Augmentation'' model, if the mid-range band is missing it performs significantly worse than the baseline. This relative decrease in performance is in agreement with the work of Hermann et al. \cite{hermann2019origins}, who noted that models trained with data augmentation tend to rely on image textures -- which occupy the mid-range frequency bands. 

On the other hand, our anti-aliased model obtained a consistent improvement over the baseline model in all of the 16 bins. This effect is further boosted when combined with data augmentation. Similar results were also observed with 8 and 32 bins.
 \autoref{appendix:smooth} extends these experiments to our models using smooth activation functions.
 We show that their benefit 
 is complementary and combining them  with anti-aliasing produces the best results.

{\setlength{\tabcolsep}{0.1em}
\begin{table*}[ht!]
\scriptsize
\begin{center}
\begin{tabular}{@{}lccccccccccccccccccccc@{}}
  \toprule
  & 
  \multicolumn{3}{c}{Noise} & \phantom{a} & \multicolumn{4}{c}{Blur} & \phantom{a} & \multicolumn{4}{c}{Weather} & \phantom{a} & \multicolumn{4}{c}{Digital} & mCE & Clean err. \\
  \cmidrule{2-4} \cmidrule{6-9} \cmidrule{11-14} \cmidrule{16-19}
  Method  
  & Gauss. & Shot & Impulse && Defocus & Glass & Motion & Zoom && Snow & Frost & Fog & Bright && Contrast & Elastic & Pixel & JPEG && \\
  \midrule
  \DTLforeach{data_aug_corruptions}{%
    \method=Method,%
    \cleanerr=CLEAN,%
    \gausserr=GAUSS,%
    \shoterr=SHOT,%
    \impulseerr=IMPULSE,%
    \defocuserr=DEFOCUS,%
    \glasserr=GLASS,%
    \motionerr=MOTION,%
    \zoomerr=ZOOM,%
    \snowerr=SNOW,%
    \frosterr=FROST,%
    \fogerr=FOG,%
    \brighterr=BRIGHT,%
    \contrasterr=CONTRAST,%
    \elasticerr=ELASTIC,%
    \pixelerr=PIXEL,%
    \jpegerr=JPEG,%
    \mce=MCE%
  }{%
    \ifthenelse{\value{DTLrowi}=1}{%
    }{%
      \ifthenelse{\value{DTLrowi}=4 \OR \value{DTLrowi}=10}{\\\midrule}{\\}%
    }%
    \dtlformat{\method}&%
    \dtlformat{\gausserr}&%
    \dtlformat{\shoterr}&%
    \dtlformat{\impulseerr}&&%
    \dtlformat{\defocuserr}&%
    \dtlformat{\glasserr}&%
    \dtlformat{\motionerr}&%
    \dtlformat{\zoomerr}&&%
    \dtlformat{\snowerr}&%
    \dtlformat{\frosterr}&%
    \dtlformat{\fogerr}&%
    \dtlformat{\brighterr}&&%
    \dtlformat{\contrasterr}&%
    \dtlformat{\elasticerr}&%
    \dtlformat{\pixelerr}&%
    \dtlformat{\jpegerr}&%
    \dtlformat{\mce}&%
    \dtlformat{\cleanerr}%
  }
  \\ \bottomrule
\end{tabular}
\end{center}
\caption{\label{table:data_aug_corruptions} Corruption Error (CE), mCE, and Clean Error values when including our anti-aliasing variations on top of ResNet-50.
Adding anti-aliasing leads to a lower error (mCE) than all existing models with the exception of ANT. ANT uses adversarial training and has an extra generative network, is significantly more expensive to train, has a higher clean error and only performs better than ours in 6 of the 15 corruptions.
Our model uses fewer and smaller filters than \cite{zhang2019shiftinvar} but at precise locations.
 Models with data augmentation were trained for longer. For all columns, lower is better. }
\end{table*}
}
\begin{table}[ht]
\scriptsize
\begin{center}
\begin{tabular}{@{}lc@{}}
  \toprule
  Method & Average rank \\
  \midrule
  RelationNet~\cite{triantafillou2020metadataset}   & 9.80 \\
  k-NN~\cite{triantafillou2020metadataset}          & 8.95 \\
  MatchingNet~\cite{triantafillou2020metadataset}   & 8.60 \\
  fo-MAML~\cite{triantafillou2020metadataset}       & 8.15 \\
  Finetune~\cite{triantafillou2020metadataset}      & 7.10 \\
  ProtoNet~\cite{triantafillou2020metadataset}      & 6.70 \\
  fo-Proto-MAML~\cite{triantafillou2020metadataset} & 4.80 \\
  CNAPs~\cite{requeima2019fast}                     & 4.45 \\
  SUR~\cite{dvornik2020selecting}                   & 3.40 \\
  URT~\cite{liu2021universal}                       & 2.40 \\
  \midrule
  SUR + Anti-aliased + GELU (ours) & {\bf 1.65} \\
  \bottomrule
\end{tabular}
\end{center}
\caption{\label{table:md_rank} Average rank over all Meta-Dataset test sources for approaches that learn on all Meta-Dataset training sources. We recompute ranks after including our proposed approach into Meta-Dataset's public leaderboard.}
\end{table}


\subsection{ImageNet-C}
\label{sec:imagenetc}


\autoref{table:data_aug_corruptions} presents a comparison of our models on the \ImNetC leaderboard. 
It uses the corruption error (CE) measurement proposed by \cite{hendrycks2019robustness} that is defined as $ CE^{f}_{c} = \frac{\sum^{5}_{s=1} E^{f}_{s,c}}{\sum^{5}_{s=1} E^{AlexNet}_{s,c}}$ where $E^{f}_{s,c}$ is the top-1 error of a classifier $f$ for a corruption $c$ with severity $s$. The mean Corruption Error (mCE) is taken by averaging over all the 15 corruptions. 
The `Clean err.'' column shows the classifier's top-1 error on the original \ImNetClean, included in the table for contrasting gains on clean and corrupted test sets.


%

Our anti-aliased model, described in \autoref{sec:imagenet_results}, achieves mCE of 70.0\% on \ImNetC and 22.5\%  ``Clean err.'' on \ImNetClean, which represent a significant improvement over the results reported in \cite{zhang2019shiftinvar}. Also note that our model uses fewer and smaller filters than \cite{zhang2019shiftinvar}
but at key positions discussed in \autoref{sec:methods}. Zhang's best model uses 7x7 filters at every sub-sampling operation. \autoref{appendix:ood_generalization} presents a direct comparison to \cite{zhang2019shiftinvar}.

Next we show that performance can be further improved by the use of data-augmentation and smooth activation functions (Swish \cite{ramachandran2017searching}). These models were trained for longer (180 epochs) in order to replicate training conditions of the remaining references mentioned in the table. Similar to \cite{hendrycks2020augmix}, our data-augmented models do not use augmentations such as contrast, color, brightness, sharpness, as they overlap with the \ImNetC test set corruptions. We achieve a ``Clean'' top-1 error of 21.2\% and an mCE of 64.9\%. Note that unlike \cite{rusak2020simple} we demonstrate that mCE can be improved \emph{without sacrificing} clean error.

\subsection{Meta-Dataset with SUR}
\label{sec:meta-dataset}

Meta-Dataset is a challenging few-shot image classification benchmark which samples heterogeneous learning episodes from a diverse collection of datasets, two of which (Traffic Signs, MSCOCO) are used exclusively for evaluation and are therefore considered out-of-domain. We apply anti-aliasing and smooth-activation (GELU) to SUR~\cite{dvornik2020selecting}, which is a competitive few-shot classification approach that trains one feature extractor per training domain and combines their representations during inference. We obtain SoTA average rank across all test datasets among approaches which learn from all available training classes (\autoref{table:md_rank}). Additional experimental details and accuracy breakdowns are presented in \autoref{appendix:metadataset}.

\section{Conclusion}
Drawing from the classical sampling theorem from signal processing, we proposed simple architectural improvements to convolutional architectures to counter aliasing occurring at various stages. These changes lead to substantial performance gains on both i.i.d. and o.o.d. generalization, 
and were shown to boost the impact of data augmentation and smooth activation functions by extending their effect across the spectrum. Compared to other performance enhancement techniques, anti-aliasing is simple to implement, computationally inexpensive, and does not require additional trainable parameters. In all our experiments, we could not find a setting where it degraded the performance. Therefore, we recommend their use as a standard component of convolutional architectures.

\clearpage
{\small
\bibliographystyle{ieee_fullname}
\bibliography{iccv}
}

\newpage .
\newpage
\appendix

\section{The power spectral density of Binomial filters}
\label{appendix:DFT}

\begin{figure}
\centering
\includegraphics[width=1.0\linewidth]{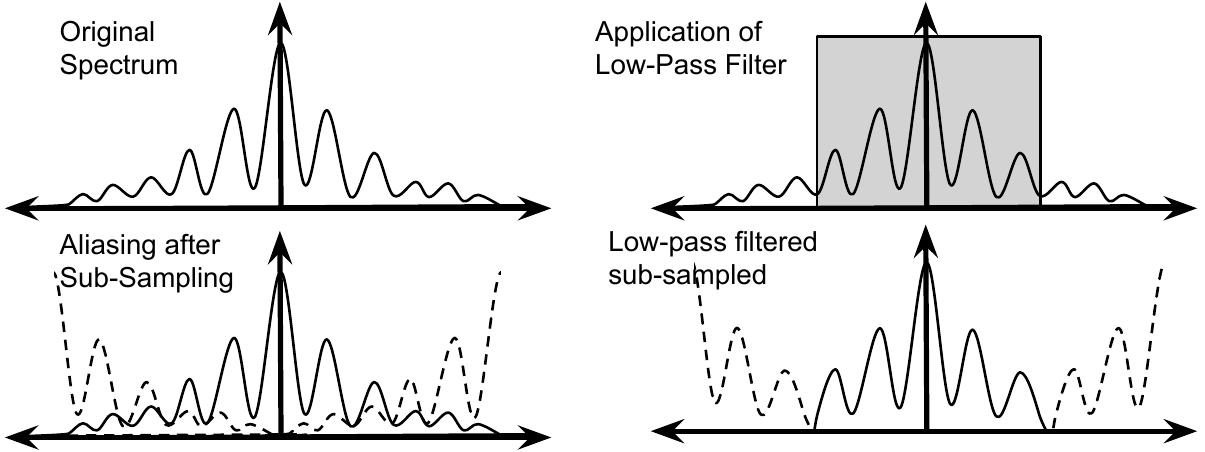}
\caption{Subsampling without low-pass filtering causes the spectra to overlap and become corrupted. Left: after subsampling spectra could overlap, which is called aliasing; Right: subsampling preceded by low-pass filtering with an ideal low-pass filter prevents corruption.}
\label{fig:aliasing_theory}
\end{figure}

This section reiterates the Fourier transform and its properties, and we provide the Fourier transforms for the filters used in our anti aliasing method. For a full exposition to the Fourier transform and frequency analysis please see \cite{openheim1997signals}. In general the Fourier transform has both magnitude and phase components, our discussion focuses on the magnitude which is referred to as the power spectral density.

Figure \ref{fig:aliasing_theory} depicts the basic theory of frequency aliasing. When a signal is subsampled it's spectrum is replicated at distances inversely proportional to the sampling rate. Frequencies from these copies additively spill into the original signal, corrupting the original frequency components (``Aliasing after Sub-Sampling''). This aliasing can be prevented by the application of a low-pass filter, by which the lower-frequency components of the original signal can be preserved. 

We use binomial filters as a low-pass filter because of their finite support size. Here we discuss the power spectral density of the 1D filter, which could be extended to the 2D case by an outer product. Examples of discrete binomial filters include $[1, 2, 1]$, $[1, 4, 6, 4, 1]$, etc, which can be generated from Pascal's triangle. Using the bracket $[\cdot]$ to index the signal, $x$, and defining the discrete Dirac delta function as $\delta$, the filters of interest in our work are defined as:

\begin{align*}
    x_1[n] &= \delta[n-1] + 2\delta[n] + \delta[n+1] \\
    x_2[n] &= \delta[n-2] + 4\delta[n-1] + 6\delta[n] +  4\delta[n+1] + \\ & \delta[n+2] \\
    x_3[n] &= \delta[n-3] + 6\delta[n-2] + 15\delta[n-1] + 20\delta[n] + \\
           & 15\delta[n+1] + 6\delta[n+2] + \delta[n+3] \\
\end{align*}

Using the Fourier identities from Equations~\ref{fourier}, we obtain the the corresponding signals in Fourier domain, denoting the angular frequency with $w$:

\begin{align*}
    x_1[w] &= 2 + 2\cos(w) \\
    x_2[w] &= 6 + 8 \cos(w) + 2 \cos(w) \\
    x_3[w] &= 20 + 30 \cos(w) + 12 \cos(2w) + \cos(3w) \\
\end{align*}

\begin{figure}[ht]
    \centering
    \includegraphics[width=\linewidth]{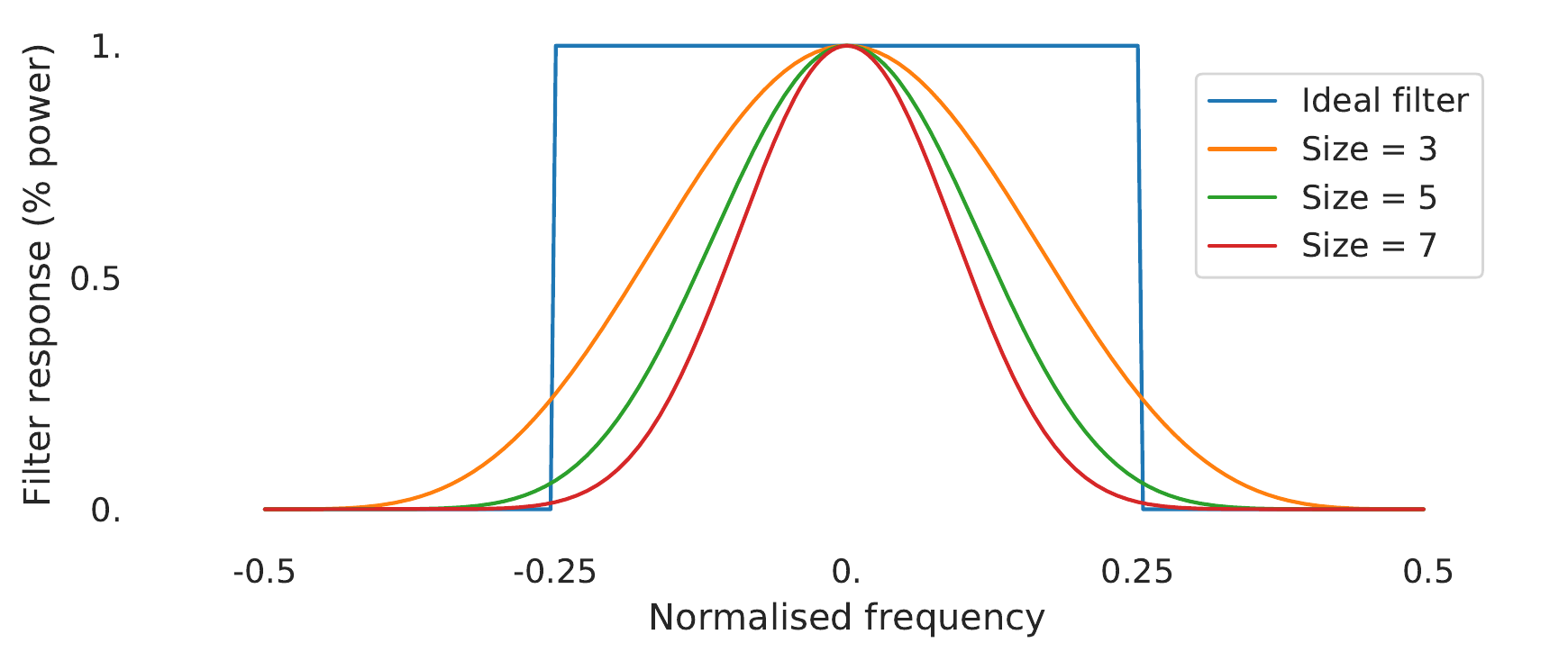}
    \caption{Power spectral density for filters used for anti-aliasing. The blue curve indicates an ideal filter for a subsampling with stride 2. However, as an ideal filter is too computationally expensive, we plot three alternatives of varying support size.}
    \label{fig:filter_psd}
\end{figure}

\autoref{fig:filter_psd} shows the power spectral densities (magnitude of Fourier transform). This diagram highlights the trade-offs. A filter with larger support size, $k=7$, attenuates more power at the cut-off frequency. However, more of the frequencies just below the cut-off frequency are also attenuated. A filter with smaller support size, $k=3$, attenuates less power at the cut-off frequency, but maintains more information from the high frequencies just below the cut-off frequency.

\begin{subequations}\label{fourier}
\begin{empheq}[box=\widefbox]{align}
\sum_{w=-\infty}^\infty x[n] e^{-jwn} &= x[w]   \\
\delta[n] &= 1 \\
\delta[n-n_0] &= e^{-jwn_0} \\
\delta[n-k] + \delta[n+k] &= 2 \cos(wk)
\end{empheq}
\end{subequations}

\section{On Smooth activations}

The optimal placement criteria presented in \autoref{sec:confounding_effects}
argues in favor of preserving the information encoded in high frequencies through layers that do not cause aliasing. Next, in \autoref{sec:arch_review}, our definition of the aliasing critical path claimed that the activation function non-linearity may produce high-frequency content. These arguments sustain the placement of low-pass filters in our models and its relation to activation functions (point-wise non-linearities) present in the original architecture. 

\label{appendix:smooth}
\begin{figure}[ht]
\centering 
\begin{minipage}{\linewidth}
\centering
\subfloat[\footnotesize x]
{\includegraphics[width=0.49\textwidth]{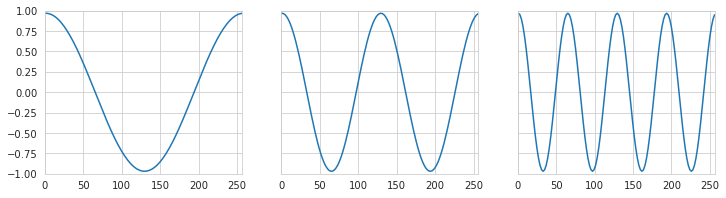}}\hfill
\subfloat[\footnotesize $\mathcal{F}(x)$ ]{\includegraphics[width=0.49\textwidth]{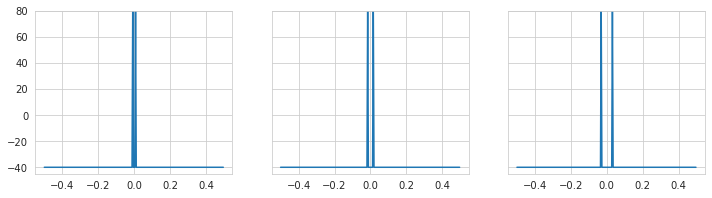}} \\
\subfloat[\footnotesize $ReLU(x)$]
{\includegraphics[width=0.49\textwidth]{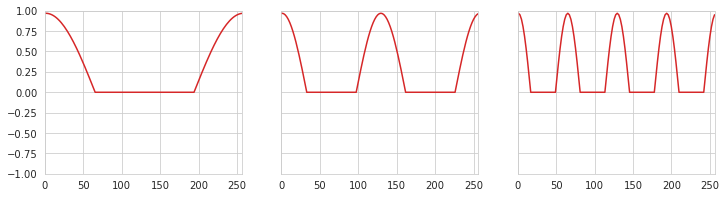}}\hfill
\subfloat[\footnotesize $\mathcal{F}(ReLU(x))$ ]{\includegraphics[width=0.49\textwidth]{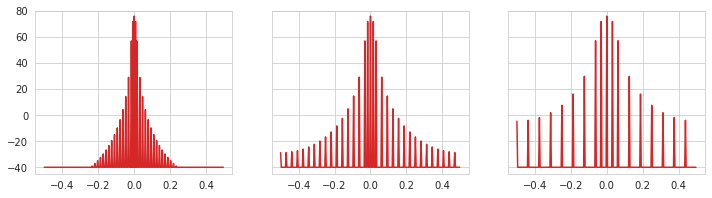}} \\
\subfloat[\footnotesize $GeLU(x)$]
{\includegraphics[width=0.49\textwidth]{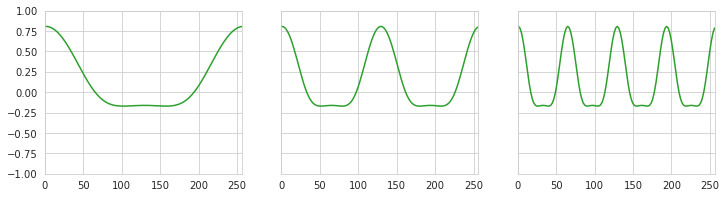}}\hfill
\subfloat[\footnotesize $\mathcal{F}(GeLU(x))$ ]{\includegraphics[width=0.49\textwidth]{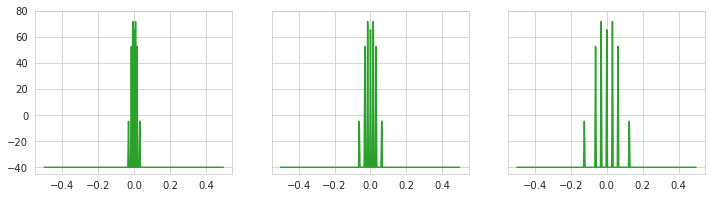}} \\

\subfloat[\footnotesize $Swish(x)$]
{\includegraphics[width=0.49\textwidth]{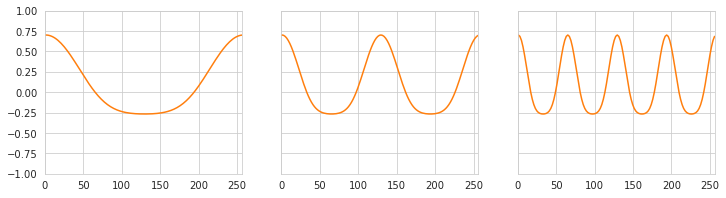}}\hfill
\subfloat[\footnotesize $\mathcal{F}(Swish(x))$ ]{\includegraphics[width=0.49\textwidth]{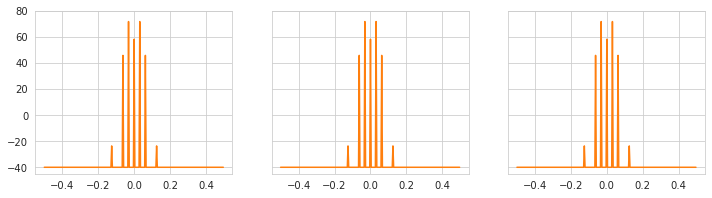}} \\
\end{minipage}

\caption{\label{fig:smooth_activation} 
Illustration of the spectrum changes caused by non-linear activation functions. First row shows three different inputs, each containing a different sinusoidal wave. 
Comparison between ReLU and smooth activations (GeLU and Swish) illustrates the impact on the spectra of the resulting signals. \emph{Left:} spatial-domain representation. \emph{Right:} frequency-domain representation.
}
\label{fig:smooth}
\end{figure}

This section illustrates the changes in spectral distribution caused by non-linear activation functions. Note that they introduce high-frequency components only if the inputs to the activation function span the non-linearity, i.e. for the activation functions considered here, inputs must be positive and negative.


Without loss of generality, Figure \ref{fig:smooth} illustrates the change in the power spectral density of 1D signals under different activation functions. The first row shows both spatial ($x$) and frequency domain ($\mathcal{F}$) representations of 1D sinusoidal waves, each containing a single frequency, i.e. this is our input signal without the application of any non-linearity. 
The following rows illustrate the spatial and frequency domain representations corresponding to the output of different non-linearities. 
The ``elbow'' spatial-domain characteristic (around zero) imposed by $ReLU$ is reflected in the resulting frequency-domain representation with the introduction of higher frequency components that were not present in the original input. The power of these new components decreases with their frequency. 
The results obtained by applying smooth-activation ($GeLU$ and $Swish$) to the same input signal also introduces new high frequency components, but with much faster decay.      
The proofs of the duality between spatial smoothness and frequency component decay can be found in \cite{katznelson_2004} (page 41 shows that smoothness in time/spatial-domain implies decay in frequency-domain).

The use of smooth activations were previously shown to improve robustness to adversarial attacks \cite{Xie_2020_SAT}. The results of the ablation studies presented in sections \ref{sec:imagenetc} and \ref{sec:meta-dataset} extend these findings to show their impact on a broader concept of o.o.d. robustness.

\begin{figure*}[htb]
\centering
\includegraphics[width=\linewidth]{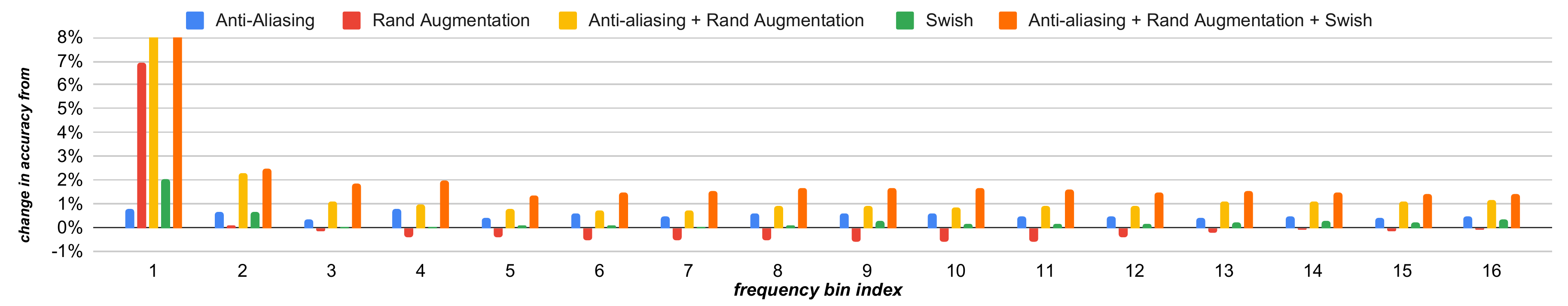}
\caption{\label{fig:robustness_to_frequency_removal_smooth} Illustration of the relative performance impact on pre-trained models when tested on images that have 1/16 of their spectral band removed.  
Our anti-aliased model performance is higher than the baseline in all spectral bands. The use of smooth activation functions alone have a larger impact in lower bands.
The combined model (AA+DA+Swish) combines the advantages of all three. Note that this figure illustrates relative improvements to the baseline results taken under the same experiment. Baseline degradation curve is presented in \autoref{fig:robustness_to_frequency_removal}}
\end{figure*}

\autoref{fig:robustness_to_frequency_removal_smooth} complements the results presented in \autoref{fig:robustness_to_frequency_removal} by including the performance gains across different spectral bands for a model trained using smooth activation (``Swish'') and also a model combining anti-aliasing with smooth activation and data augmentation (``Anti-aliasing + Rand Augmentation +Swish''). 
Similar to data-augmentation alone, smooth activations lead to improved performance mainly when the in lower bins are filtered, in contrast to the anti-aliasing which improves performance across the entire spectrum. The figure also shows that the combined model presents the best results at all frequency ranges, extending the benefits of using smooth activation functions, initially observed mainly in the lower frequencies, to the entire spectrum. 

\begin{figure*}[t]
\centering
\captionsetup[subfigure]{labelformat=empty}
\begin{minipage}{0.99\textwidth}
\centering
{\includegraphics[width=0.06\textwidth]{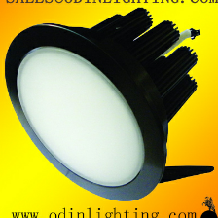}}\hfill
{\includegraphics[width=0.06\textwidth]{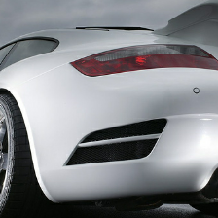}}\hfill
{\includegraphics[width=0.06\textwidth]{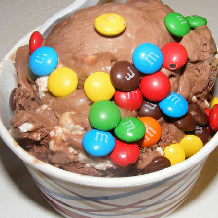}}\hfill
{\includegraphics[width=0.06\textwidth]{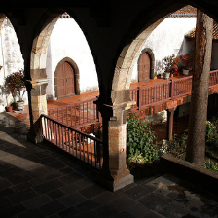}}\hfill
{\includegraphics[width=0.06\textwidth]{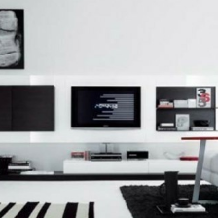}}\hfill
{\includegraphics[width=0.06\textwidth]{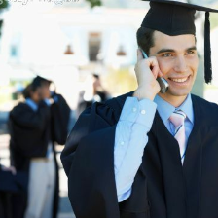}}\hfill
{\includegraphics[width=0.06\textwidth]{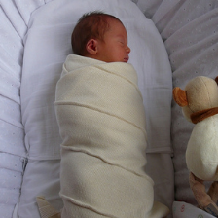}}\hfill
{\includegraphics[width=0.06\textwidth]{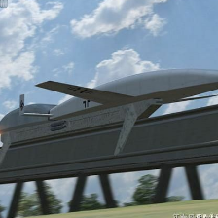}}\hfill
{\includegraphics[width=0.06\textwidth]{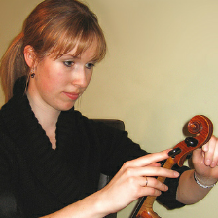}}\hfill
{\includegraphics[width=0.06\textwidth]{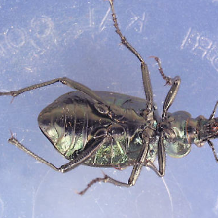}}\hfill
{\includegraphics[width=0.06\textwidth]{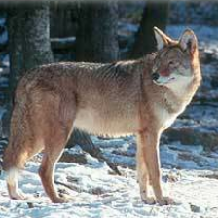}}\hfill
{\includegraphics[width=0.06\textwidth]{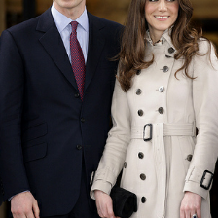}}\hfill
{\includegraphics[width=0.06\textwidth]{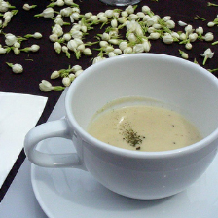}}\hfill
{\includegraphics[width=0.06\textwidth]{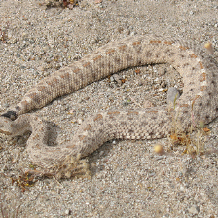}}\hfill
{\includegraphics[width=0.06\textwidth]{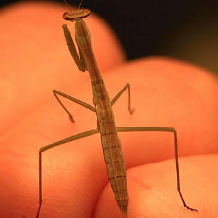}}\hfill
{\includegraphics[width=0.06\textwidth]{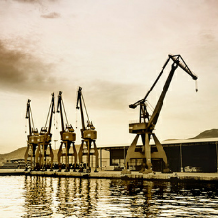}}\hfill
\subfloat[\footnotesize bin 1]
{\includegraphics[width=0.06\textwidth]{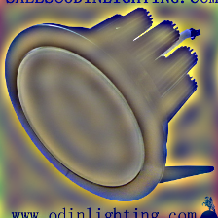}}\hfill
\subfloat[\footnotesize bin 2]
{\includegraphics[width=0.06\textwidth]{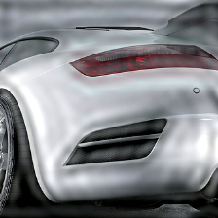}}\hfill
\subfloat[\footnotesize bin 3]
{\includegraphics[width=0.06\textwidth]{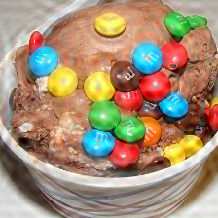}}\hfill
\subfloat[\footnotesize bin 4]
{\includegraphics[width=0.06\textwidth]{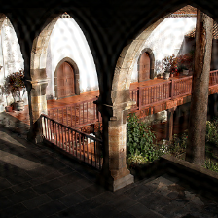}}\hfill
\subfloat[\footnotesize bin 5]
{\includegraphics[width=0.06\textwidth]{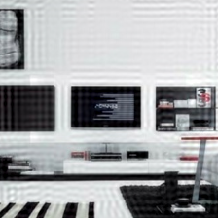}}\hfill
\subfloat[\footnotesize bin 6]
{\includegraphics[width=0.06\textwidth]{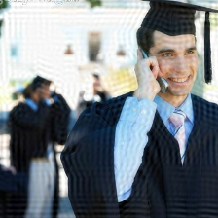}}\hfill
\subfloat[\footnotesize bin 7]
{\includegraphics[width=0.06\textwidth]{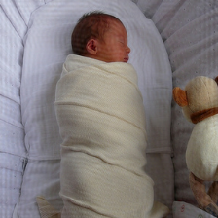}}\hfill
\subfloat[\footnotesize bin 8]
{\includegraphics[width=0.06\textwidth]{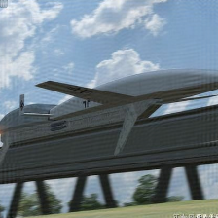}}\hfill
\subfloat[\footnotesize bin 9]
{\includegraphics[width=0.06\textwidth]{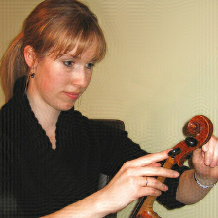}}\hfill
\subfloat[\footnotesize bin 10]
{\includegraphics[width=0.06\textwidth]{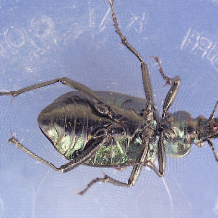}}\hfill
\subfloat[\footnotesize bin 11]
{\includegraphics[width=0.06\textwidth]{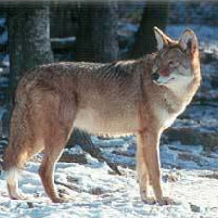}}\hfill
\subfloat[\footnotesize bin 12]
{\includegraphics[width=0.06\textwidth]{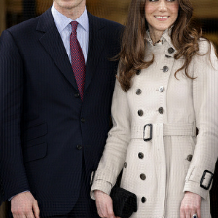}}\hfill
\subfloat[\footnotesize bin 13]
{\includegraphics[width=0.06\textwidth]{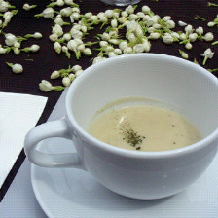}}\hfill
\subfloat[\footnotesize bin 14]
{\includegraphics[width=0.06\textwidth]{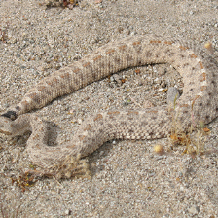}}\hfill
\subfloat[\footnotesize bin 15]
{\includegraphics[width=0.06\textwidth]{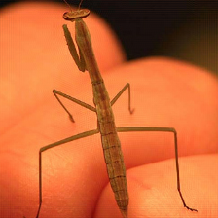}}\hfill
\subfloat[\footnotesize bin 16]
{\includegraphics[width=0.06\textwidth]{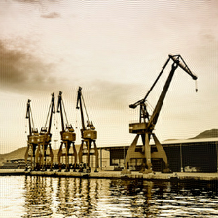}}
\end{minipage}
\caption{\label{fig:images_frequency_removal2} 
Original and filtered test images pairs to illustrate the effect of the notch filter on each of the 16 frequency intervals (in order starting from lowest band). They are wrongly classified by the baseline model and correctly classified by the data-augmented model (bin 1) or the anti-aliased model (bin 2 - bin 16), but not by both. 
}
\label{fig:frequency_removal_images}
\end{figure*}

Figure \ref{fig:frequency_removal_images} depicts a random sample of images and the corresponding images with various frequency bands removed. Note that when high frequency bands are filtered the images appear nearly indistinguishable from the originals. 


\section{EfficientNets}
\label{appendix:efficientNets}

Large EfficientNet models are constructed by expanding a baseline model (EfficientNet-B0) in terms of model depth, model width, and input image resolution. For instance, EfficientNet-B0 adopts an input image of 224$\times$224, EfficientNet-B1 scales it to 240$\times$240 pixels and EfficientNet-B7 up to 600$\times$600 pixels. 

The results presented in this section were obtained with a EfficientNet-B0 in order to contrast the impact of aliasing on ResNet with EfficientNet models without introducing confounding effects related to input resolution.
\autoref{table:efficientnet} shows that EfficientNet-B0 models also benefit from the introduction of small 
non-trainable low pass filters. This effect holds when training from \ImNetClean and using RandAugmentation.
A comparison to ResNet-50 results shows that the impact of aliasing is smaller in EfficientNet-B0 -- perhaps as a result of the fact that neural architecture search may have found an architecture which partially mitigates these effects. 
EfficientNet models critical paths have a small $3 \times 3$ filter (\autoref{fig:efficientnets}) that justifies the relative lower impact of anti-aliasing the EfficientNet model when compared to the impact observed on Resnet models. These results confirm our hypotheses that adding anti-aliasing to the critical paths is necessary in complement to the existing filters capacity (originally composed by 3 $\times$ 3 filters).

\begin{table}[ht]
\scriptsize
\begin{center}
\begin{tabular}{@{}lcc@{}}
  \toprule
  Model & Top-1 Acc. \\
  \midrule
  Resnet-50 & 76.49   \tiny $\pm$ 0.06 \\
  Resnet-50 Anti-aliased & 77.47   \tiny $\pm$ 0.12 \\
  Resnet-50 + Rand-Augmentation & 77.38 \tiny $\pm$ 0.06 \\
  Resnet-50 Anti-Aliased + Rand-Augmentation & 78.85 \tiny $\pm$ 0.02 \\
  
\midrule

  EfficientNet-B0 & 76.40 \tiny $\pm$ 0.01\\
  EfficientNet-B0 Anti-aliased (k=3)& 76.65 \tiny $\pm$ 0.12\\
  EfficientNet-B0 Anti-aliased (k=5)&  76.58 \tiny $\pm$ 0.07\\

\midrule

  EfficientNet-B0 + Rand-Augmentation& 76.98	\tiny 		$\pm$0.14\\
  EfficientNet-B0 Anti-Aliased (k=3) + Rand-Augmentation & 77.17 \tiny $\pm$0.10 \\
  EfficientNet-B0 Anti-Aliased (k=5) + Rand-Augmentation &  77.12\tiny $\pm$ 0.08	\\
  \bottomrule
\end{tabular}
\end{center}
\caption{\label{table:efficientnet} Comparison of the impact of aliasing on top-1 accuracy of EfficientNet versus Resnet models for input image with resolution $224\times224$. Table contains corresponding baselines, our anti-aliased versions and their combinations with data-augmentation. The values correspond to the mean accuracy over 3 runs with different seeds.
Anti-aliased models present the best performance, but impact on Resnet-50 is larger than on EfficientNet-B0. 
Resnet-50 contains aliasing critical paths that lack a minimum size to represent low-pass filters, while EfficientNet-B0 don't have such critical bottlenecks.  
}
\end{table}



\begin{figure}[t]
\centering
\captionsetup[subfigure]{labelformat=empty}
\begin{minipage}{0.4\textwidth}
\centering
\includegraphics[width=\textwidth]{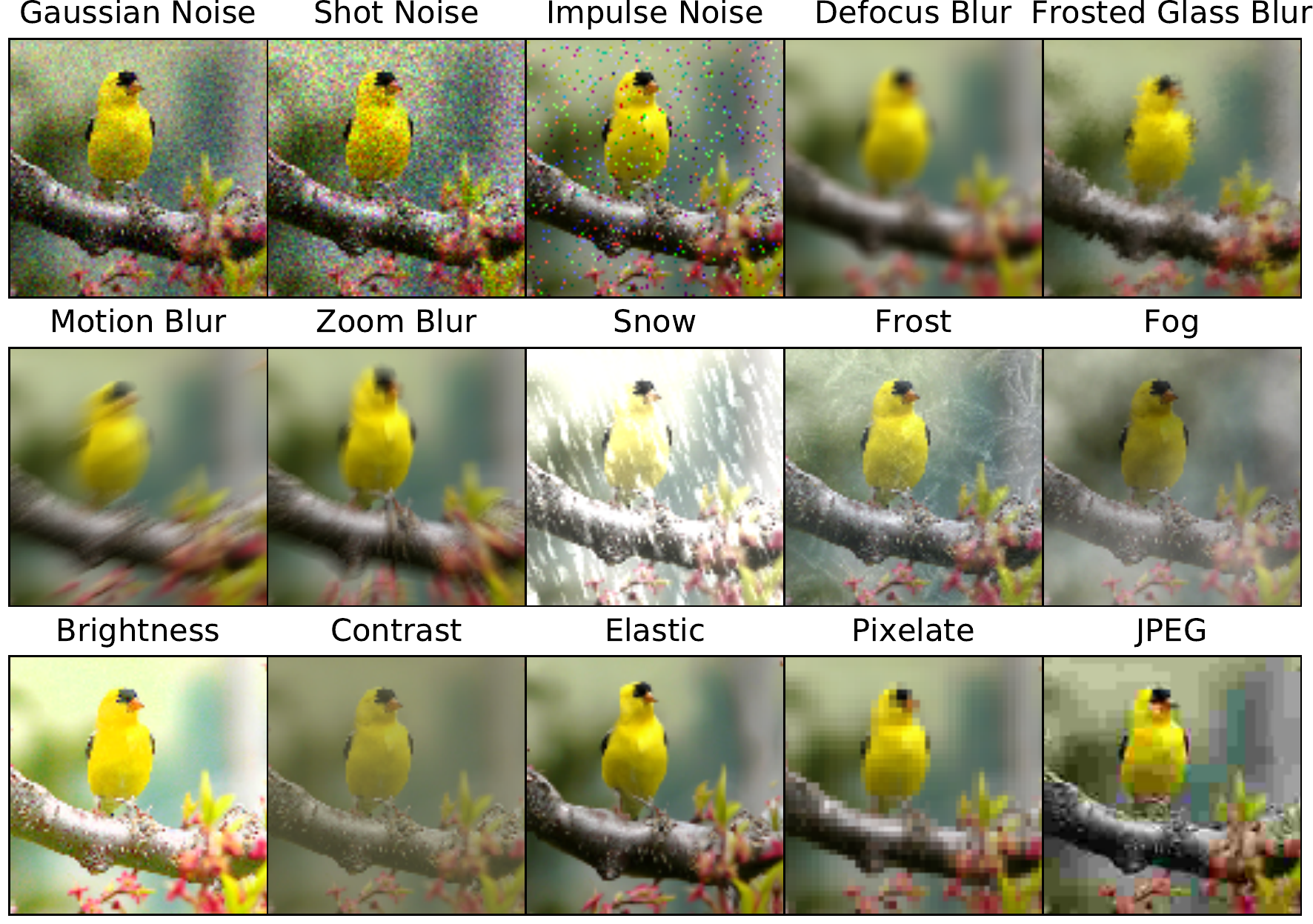}
\end{minipage}
\caption{\label{fig:datasets_imagenetC}{\em Left}:  \ImNetC corruptions. {\em Right}: Samples from all 10 data sources included in the Meta-Dataset benchmark.
Figure from  Hendrycks and Dietterich~\cite{hendrycks2019robustness}.}
\end{figure}


{\setlength{\tabcolsep}{0.1em}
\begin{table*}[ht]
\scriptsize
\begin{center}
\begin{tabular}{@{}lcccccccccccccccccccccccc@{}}
  \toprule
  & \multicolumn{3}{c}{Blur filter location} & \phantom{a} & \multicolumn{3}{c}{Noise} & \phantom{a} & \multicolumn{4}{c}{Blur} & \phantom{a} & \multicolumn{4}{c}{Weather} & \phantom{a} & \multicolumn{4}{c}{Digital} & mCE & Clean err.\\
  \cmidrule{2-4} \cmidrule{6-8} \cmidrule{10-13} \cmidrule{15-18} \cmidrule{20-23}
  Method & skip & max-pool & block-conv && Gauss. & Shot & Impulse && Defocus & Glass & Motion & Zoom && Snow & Frost & Fog & Bright && Contrast & Elastic & Pixel & JPEG && \\
  \midrule
  \DTLforeach{clean_corrupt_error}{%
    \method=Method,%
    \isskip=SKIP,%
    \ismaxpool=MAX-POOL,%
    \isblockconv=BLOCK-CONV,%
    \cleanerr=CLEAN,%
    \gausserr=GAUSS,%
    \shoterr=SHOT,%
    \impulseerr=IMPULSE,%
    \defocuserr=DEFOCUS,%
    \glasserr=GLASS,%
    \motionerr=MOTION,%
    \zoomerr=ZOOM,%
    \snowerr=SNOW,%
    \frosterr=FROST,%
    \fogerr=FOG,%
    \brighterr=BRIGHT,%
    \contrasterr=CONTRAST,%
    \elasticerr=ELASTIC,%
    \pixelerr=PIXEL,%
    \jpegerr=JPEG,%
    \mce=MCE%
  }{%
    \ifthenelse{\value{DTLrowi}=1}{%
    }{%
      \ifthenelse{\value{DTLrowi}=3}{\\\midrule}{\\}%
    }%
    \dtlformat{\method}&%
    \dtlformat{\isskip}&%
    \dtlformat{\ismaxpool}&%
    \dtlformat{\isblockconv}&&%
    \dtlformat{\gausserr}&%
    \dtlformat{\shoterr}&%
    \dtlformat{\impulseerr}&&%
    \dtlformat{\defocuserr}&%
    \dtlformat{\glasserr}&%
    \dtlformat{\motionerr}&%
    \dtlformat{\zoomerr}&&%
    \dtlformat{\snowerr}&%
    \dtlformat{\frosterr}&%
    \dtlformat{\fogerr}&%
    \dtlformat{\brighterr}&&%
    \dtlformat{\contrasterr}&%
    \dtlformat{\elasticerr}&%
    \dtlformat{\pixelerr}&%
    \dtlformat{\jpegerr}&%
    \dtlformat{\mce}&%
    \dtlformat{\cleanerr}%
  }
  \\ \bottomrule
\end{tabular}
\end{center}
\caption{\label{table:clean_corruption_error} Corruption Error (CE) on Imagenet-C corruptions, mCE, and Clean Error values when including our anti-aliasing variations. ResNet-50 and training for 90 epochs. Lower is better. We see that adding anti-aliasing decreases the errors on all corruptions except for Pixel and Blur. The errors were computed on the model achieving the median performance on ImageNet across 3 seeds. In our models, blur is not applied at the first convolutional layer (due to its large spatial support) and on the other sub-sampled modules it is applied at the precise location sustained by spectral analysis, ie. at the sub-sampling operation before its non linearities, as opposed to after as in \cite{zhang2019shiftinvar}.}
\end{table*}}
{\setlength{\tabcolsep}{0.1em}
\begin{table*}[ht]
\scriptsize
\begin{center}
\begin{tabular}{@{}lcccccccccccccccccccccccc@{}}
  \toprule
  & \multicolumn{3}{c}{Blur filter placement} & \phantom{a} & \multicolumn{3}{c}{Noise} & \phantom{a} & \multicolumn{4}{c}{Blur} & \phantom{a} & \multicolumn{4}{c}{Weather} & \phantom{a} & \multicolumn{4}{c}{Digital} & mCE & Clean err. \\
  \cmidrule{2-4} \cmidrule{6-8} \cmidrule{10-13} \cmidrule{15-18} \cmidrule{20-23}
  Method & skip & max-pool & block-conv && Gauss. & Shot & Impulse && Defocus & Glass & Motion & Zoom && Snow & Frost & Fog & Bright && Contrast & Elastic & Pixel & JPEG && \\
  \midrule
  \DTLforeach{imagenetc_our_variations}{%
    \method=Method,%
    \isskip=SKIP,%
    \ismaxpool=MAX-POOL,%
    \isblockconv=BLOCK-CONV,%
    \cleanerr=CLEAN,%
    \gausserr=GAUSS,%
    \shoterr=SHOT,%
    \impulseerr=IMPULSE,%
    \defocuserr=DEFOCUS,%
    \glasserr=GLASS,%
    \motionerr=MOTION,%
    \zoomerr=ZOOM,%
    \snowerr=SNOW,%
    \frosterr=FROST,%
    \fogerr=FOG,%
    \brighterr=BRIGHT,%
    \contrasterr=CONTRAST,%
    \elasticerr=ELASTIC,%
    \pixelerr=PIXEL,%
    \jpegerr=JPEG,%
    \mce=MCE%
  }{%
    \ifthenelse{\value{DTLrowi}=1}{%
    }{%
      \ifthenelse{\value{DTLrowi}=7 \OR \value{DTLrowi}=13}{\\\midrule}{\\}%
    }%
    \dtlformat{\method}&%
    \dtlformat{\isskip}&%
    \dtlformat{\ismaxpool}&%
    \dtlformat{\isblockconv}&&%
    \dtlformat{\gausserr}&%
    \dtlformat{\shoterr}&%
    \dtlformat{\impulseerr}&&%
    \dtlformat{\defocuserr}&%
    \dtlformat{\glasserr}&%
    \dtlformat{\motionerr}&%
    \dtlformat{\zoomerr}&&%
    \dtlformat{\snowerr}&%
    \dtlformat{\frosterr}&%
    \dtlformat{\fogerr}&%
    \dtlformat{\brighterr}&&%
    \dtlformat{\contrasterr}&%
    \dtlformat{\elasticerr}&%
    \dtlformat{\pixelerr}&%
    \dtlformat{\jpegerr}&%
    \dtlformat{\mce}&%
    \dtlformat{\cleanerr}%
  }
  \\ \bottomrule
\end{tabular}
\end{center}
\caption{\label{table:imagenetC_variations} Corruption Error (CE), mCE, and Clean Error values when including our anti-aliasing variations on top of ResNet-50 and training for 180 epochs with data augmentation. Adding anti-aliasing leads to a lower error than all existing models with the exception of ANT. ANT uses adversarial training and has an extra generative network, is significantly more expensive to train, has a higher clean error and has comparable Corruption Error to our simple modification. The errors were computed on the model achieving the median performance on ImageNet across 3 seeds. In our models, anti-aliasing is applied before the non linearities, as opposed to after as in \cite{zhang2019shiftinvar}.}
\end{table*}
}

\section{ImageNet-C: Robustness to Natural Corruptions}
\label{appendix:imagenetc}

\ImNetC \cite{hendrycks2019robustness} is a dataset used for evaluating the robustness of classifiers under natural image corruptions. It consists of the \ImNetClean validation set corrupted with 15 (plus four optional) types of natural corruptions under various severity levels (\autoref{fig:datasets_imagenetC} depicts \ImNetC examples). These corruptions distort the distribution of the image spectra to varying degrees. In contrast to previously proposed methods \cite{hendrycks2019robustness, hendrycks2020augmix, rusak2020simple}, which achieve increased robustness (o.o.d.) at the cost of reducing i.i.d. performance (i.e. \ImNetClean validation without corruptions), we demonstrate that our method is the first to achieve state-of-the-art robustness without compromising accuracy on i.i.d performance. 
We show that our proposed architecture is complementary to data-augmentation and helps to achieve new state-of-the-art results on \ImNetC. This result also suggests that anti-aliasing cannot be fully learned using existing augmentation strategies alone, without further architecture modifications.  

\autoref{table:clean_corruption_error}  further detail the comparison between our anti-aliased model and Zhang's model \cite{zhang2019shiftinvar} on \ImNetC, when trained under the same number of epochs. It depicts  the impact of aliasing in each of our model components (from \autoref{fig:closer_look}). Note that anti-aliasing the strided-skip connections alone already surpass Zhang's results in both \ImNetClean and \ImNetC. Our combined model further improves the results using fewer and smaller filters.

Table \autoref{table:imagenetC_variations} confirms the complementary impact of anti-alias in relation to data-augmentation and smooth activation functions in this o.o.d. setting. It contains baselines obtained with data-augmentation and/or the use of smooth activation functions alone, and contrasts them with the result obtained when also combining our anti-aliased model.
The combination of the three improved both \ImNetClean and \ImNetC results.




\section{Additional experiments with Out-of-distribution Generalization}
\label{appendix:ood_generalization}


Here we report results on two additional o.o.d. generalization tasks for the models trained with anti aliasing filters. In Section~\ref{sec:imagenetc} we analysed the robustness on \ImNetC, which consists of synthetic perturbations. Here we make a 1-to-1 comparison with~\cite{zhang2019shiftinvar} on two datasets which represent natural robustness:

\begin{enumerate}
   \item \ImNetVV~\cite{DBLP:conf/icml/RechtRSS19} are images similar to those found in the \ImNetClean dataset. However this version was collected again in 2019. A high accuracy on this dataset indicates a better generalization to the new collection policy of \ImNetVV, despite the original authors showing lower accuracy of most models on this dataset.
   \item \ImNetR~\cite{DBLP:journals/corr/abs-2006-16241} are renditions of the \ImNetClean classes, but in different styles such as sketches, paintings, or sculpture. Higher accuracy on this dataset indicates robustness to rendering method and image style. 
\end{enumerate}

In order to replicate \cite{zhang2019shiftinvar}'s pipeline we also trained our models for 90 epochs. As discussed in Section \ref{sec:questions}, our models prevent aliasing in the feature maps of the neural networks, we expect our models to generalize better to these out-of-distribution datasets. Table~\ref{tab:ood_generalization} shows our results. The table shows that our anti aliased models have better o.o.d. generalization than~\cite{zhang2019shiftinvar}, while maintaining similar accuracy on \ImNetClean. Specifically, anti aliasing improves from 24.2 to 25.2 on \ImNetR, while \cite{zhang2019shiftinvar} scores 24.1. We speculate this is due to many high frequency patterns in the renditions of \ImNetR. Secondly, on \ImNetVV the baseline is 64.6, \cite{zhang2019shiftinvar} scores 65.0, and our anti aliased model has the best accuracy at 65.2. While not the focus of our work, we show that the anti-aliasing approach shows higher accuracies on natural robustness. 

\begin{table}[t]
    \centering
    \footnotesize{
     \begin{tabular}{l|ccc} 
     \toprule
Model & ImageNet & ImageNet-R & ImageNet-V2 \\ 
\midrule
Baseline                              &    76.7 &     24.4 &     64.6 \\
Zhang '19~\cite{zhang2019shiftinvar}  &    77.2 &     24.1 &     65.0 \\
Anti aliased (ours)                   & \B{77.5} & \B{25.2} & \B{65.2} \\
     \bottomrule
     \end{tabular}
     }
    \caption{Out of distribution generalization of anti aliased models compared to~\cite{zhang2019shiftinvar}} \label{tab:ood_generalization}
\end{table}

\begin{figure}
    \centering
    \includegraphics[width=\linewidth]{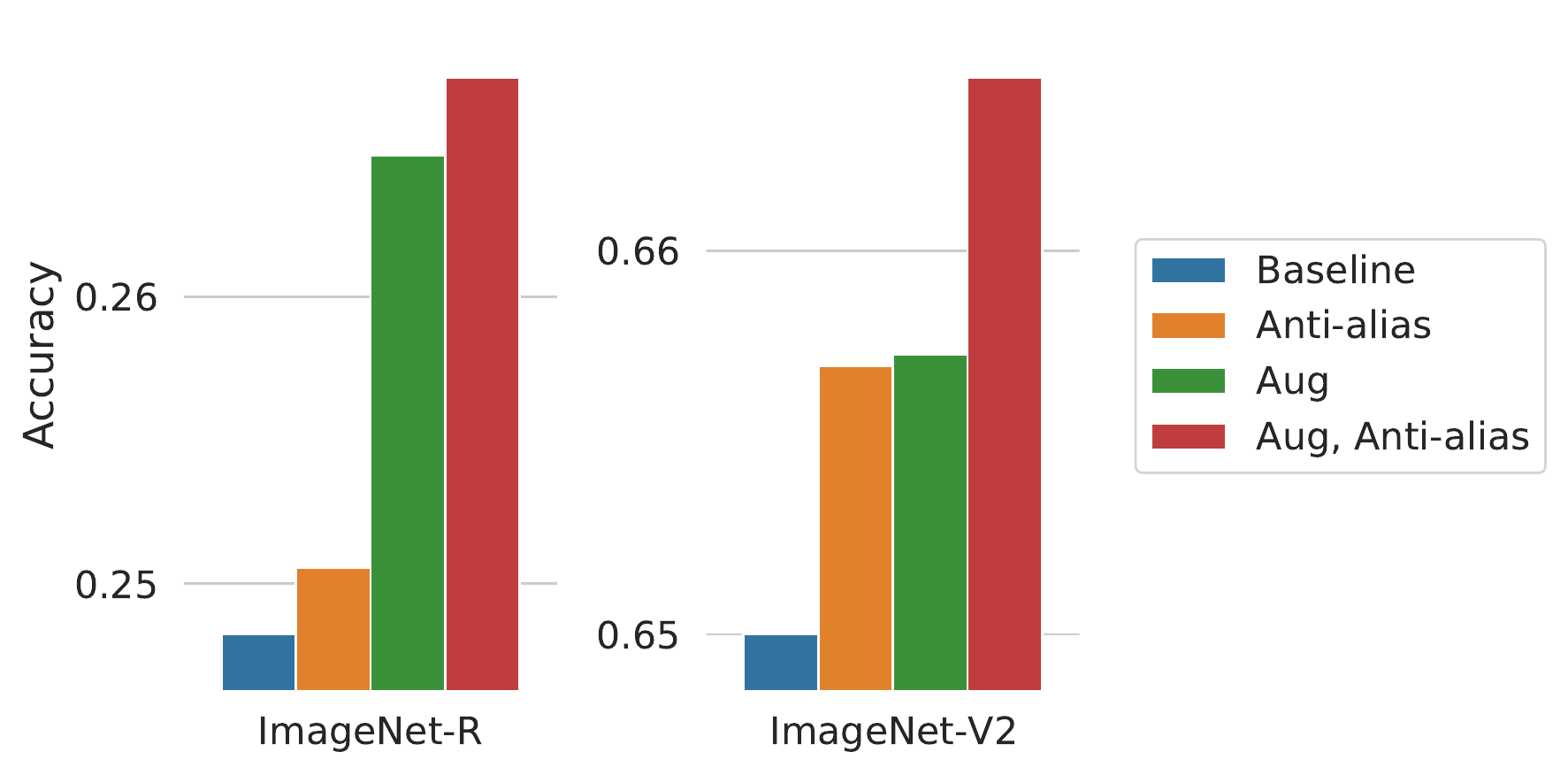}
    \caption{Our anti aliased models evaluated on two additional o.o.d. generalization tasks (described in~\autoref{appendix:ood_generalization}. Higher accuracy indicates better o.o.d. generalization. On all four datasets, we observe a complementary benefit of data augmentation with our anti aliasing method. \textit{Aug} refers to a model trained with data augmentation.}
    \label{fig:ood_generalization_two}
\end{figure}

Secondly, we analyse the complementary effect of anti aliasing and data augmentation on o.o.d. generalization. In section~\ref{sec:imagenetc} we observed a complementary effect of anti aliasing and training with data augmentation. \autoref{fig:ood_generalization_two} shows the same models evaluated on \ImNetR and \ImNetVV. For both datasets, our proposed anti aliasing method brings an improvement over the baseline. Moreover, combined with data augmentation, the increase in accuracy (and thus robustness) is higher than the improvements of either method alone.

\section{Few-shot classification, Meta-Dataset, and SUR}
\label{appendix:metadataset}

\begin{figure}[t]
\centering
\captionsetup[subfigure]{labelformat=empty}
\begin{minipage}{0.99\linewidth}
\centering
\subfloat[\footnotesize ImageNet]{\includegraphics[width=0.19\textwidth]{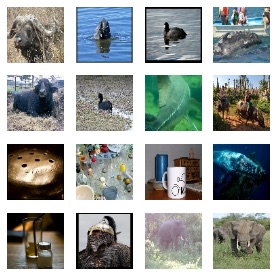}}\hfill
\subfloat[\footnotesize Omniglot]{\includegraphics[width=0.19\textwidth]{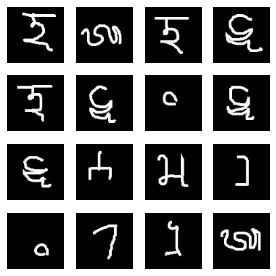}}\hfill
\subfloat[\footnotesize Aircraft]{\includegraphics[width=0.19\textwidth]{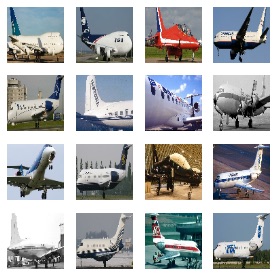}}\hfill
\subfloat[\footnotesize Birds]{\includegraphics[width=0.19\textwidth]{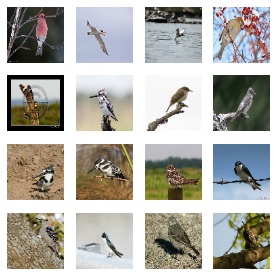}}\hfill
\subfloat[\footnotesize Textures]{\includegraphics[width=0.19\textwidth]{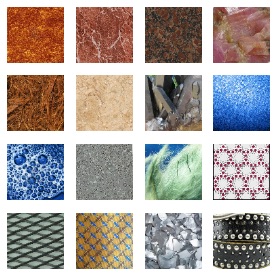}}\\
\subfloat[\footnotesize QuickDraw]{\includegraphics[width=0.19\textwidth]{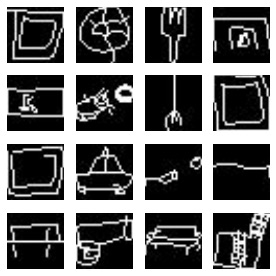}}\hfill
\subfloat[\footnotesize Fungi]{\includegraphics[width=0.19\textwidth]{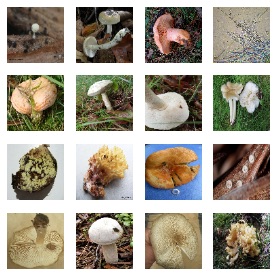}}\hfill
\subfloat[\footnotesize Flower]{\includegraphics[width=0.19\textwidth]{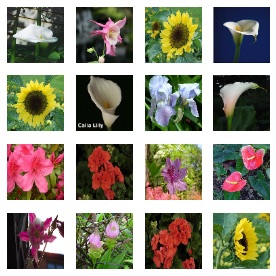}}\hfill
\subfloat[\footnotesize Traffic Signs]{\includegraphics[width=0.19\textwidth]{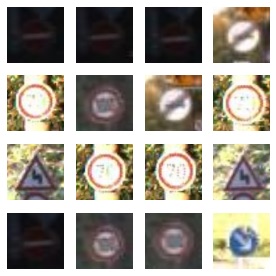}}\hfill
\subfloat[\footnotesize MSCOCO]{\includegraphics[width=0.19\textwidth]{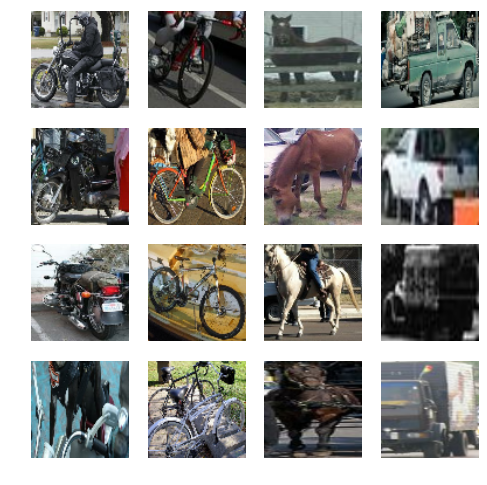}}
\end{minipage}
\caption{\label{fig:datasets_metadataset} Samples from all 10 data sources included in the Meta-Dataset benchmark.
Figures taken from  Triantafillou et al.~\cite{triantafillou2020metadataset}, respectively.}
\end{figure}

\begin{table}[ht]
\scriptsize
\begin{center}
\begin{tabular}{@{}lccccc@{}}
  \toprule
  Data source & Preprocessing & SUR* & Stride 1 \\ 
  \midrule
  \DTLforeach{meta_dataset}{%
    \source=DATASOURCE,%
    \preproc=PREPROCESSING,%
    \sur=SUR,%
    \strideone=STRIDE1%
  }{%
    \ifthenelse{\value{DTLrowi}=1}{%
        \dtlformat{\source} & \dtlformat{\preproc} &%
    }{%
      \ifthenelse{\value{DTLrowi}=9}{\\\midrule\multicolumn{2}{c}{\dtlformat{\source}} &}{\\\dtlformat{\source} & \dtlformat{\preproc} &}%
    }%
    \dtlformat{\sur}&%
    \dtlformat{\strideone}
  }
  \\\bottomrule
\end{tabular}
\end{center}

\caption{\label{table:sur_val} Subsampling in the first convolutional layer (i.e. before the first residual block) has a major impact on episodic validation performance for SUR trained on Meta-Dataset. Since the backbones are trained on $84\times84$ images, different datasets will require different amounts of upsampling or downsampling (upwards and downwards arrows in the {\em Preprocessing} column). For datasets requiring large amounts of downsampling (i.e. all datasets except Omniglot and QuickDraw), removing subsampling in the first convolutional layer ({\em Stride 1} column) shows clear benefits when compared with SUR's ResNet-18 implementation ({\em SUR*} column).
}
\end{table}
{\setlength{\tabcolsep}{0.4em}
\begin{table*}[ht]
\begin{center}
\scriptsize
\begin{tabular}{@{}lcccccccc@{}}
  \toprule
  & SUR* & 
  \multicolumn{2}{c}{Anti-aliased} & GELU & 
  \multicolumn{2}{c}{Anti-aliased + GELU} \\ 
  \cmidrule{3-4} \cmidrule{6-7}  
  Data source && $k=3$ & $k=5$ 
  && 
  $k=3$ & $k=5$
  \\
  \midrule
  \DTLforeach{meta_dataset_test}{%
    \source=DATASOURCE,%
    \sur=SUR,%
    \surs=SUR-S,%
    \aasthree=AAS3,%
    \aasthrees=AAS3-S,%
    \aasfive=AAS5,%
    \aasfives=AAS5-S,%
    \aasseven=AAS7,%
    \aassevens=AAS7-S,%
    \gelu=GELU,%
    \gelus=GELU-S,%
    \aasubgthree=AASUBG3,%
    \aasubgthrees=AASUBG3-S,%
    \aasubgfive=AASUBG5,%
    \aasubgfives=AASUBG5-S,%
    \aasubgseven=AASUBG7,%
    \aasubgsevens=AASUBG7-S,%
    \aasgthree=AASG3,%
    \aasgthrees=AASG3-S,%
    \aasgfive=AASG5,%
    \aasgfives=AASG5-S,%
    \aasgseven=AASG7,%
     \aasgsevens=AASG7-S%
  }{%
    \ifthenelse{\value{DTLrowi}=1}{%
    }{%
      \ifthenelse{\value{DTLrowi}=8 \OR \value{DTLrowi}=14}{\\\midrule}{\\}%
    }%
    \source&%
    \sur\ifthenelse{\value{DTLrowi}<14}{\tiny$\pm$\surs}{}&%
    \aasthree\ifthenelse{\value{DTLrowi}<14}{\tiny$\pm$\aasthrees}{}&%
    \aasfive\ifthenelse{\value{DTLrowi}<14}{\tiny$\pm$\aasfives}{}&%
    \gelu\ifthenelse{\value{DTLrowi}<14}{\tiny$\pm$\gelus}{}&%
    \aasgthree\ifthenelse{\value{DTLrowi}<14}{\tiny$\pm$\aasgthrees}{}&%
    \aasgfive\ifthenelse{\value{DTLrowi}<14}{\tiny$\pm$\aasgfives}{}&%
  }
  \\\bottomrule
\end{tabular}
\end{center}
\caption{\label{table:sur_test} 
Evaluation of SUR models on 600 test episodes from Meta-Dataset. Columns: \emph{SUR} shows baseline performance using original backbones.  
\emph{Anti-aliased}: shows the effect of adding blur to strided-skip connections with different blur kernel sizes ($k$), 
\emph{GELU}: replaces ReLU activations with GELU, 
\emph{Anti-aliased + GELU}: combines the anti-aliased model with GELU activations. Meta-Dataset Anti-aliased backbones used stride 1 in the first convolutional layer with input size $84\times84$. Conclusions: adding blur at skip connections improves performance with or without GELU activations. GELU actions improve performance on their own. The best result is achieved by combining blur on skip connections with GELU activations.}
\end{table*}
}


The objective behind few-shot classification is to create models which can learn on new problems with only a handful of labeled training examples. The evaluation procedure it prescribes is to form test {\em episodes} by subsampling classes from a held-out set of classes and sampling examples from those classes that are partitioned into a {\em support} (i.e. training) and a {\em query} (i.e. test) set of examples. The model is tasked with training on the support set and is evaluated on its query set accuracy, finally the query set accuracies of many test episodes are averaged to obtain a measure of model performance on new learning problems. A detailed description of the setup can be found in~\cite{triantafillou2020metadataset}.

Meta-Dataset~\cite{triantafillou2020metadataset} is a large-scale few-shot classification benchmark that was introduced as a more realistic and challenging alternative to popular benchmarks such as mini-ImageNet \cite{vinyals2016matching}. While mini-ImageNet is constructed out of ImageNet classes (using 64, 16, and 20 classes to sample training, validation, and test episodes, respectively), Meta-Dataset is constructed out of many heterogeneous datasets whose classes are themselves partitioned into training, validation, and test sets of classes. Meta-Dataset, therefore, is a more challenging dataset in terms of robustness to distribution shift, which is compounded by the fact that two of its data sources (MSCOCO and Traffic Signs) are strictly reserved for test episodes (\autoref{fig:datasets_metadataset} depicts Meta-Dataset examples).

SUR~\cite{dvornik2020selecting} tackles Meta-Dataset's domain heterogeneity by training separate backbones for each of the 8 data sources that define a training split of classes. Each backbone is trained to minimize classification error by sampling batches from its corresponding training set of classes. Hyperparameter selection is performed by evaluating on episodes sampled from each backbone's corresponding validation set of classes using a nearest centroid classifier (NCC) on top of the backbone embedding. Finally, during testing, all backbone embeddings are individually gated and concatenated to form a single representation used by a NCC. An optimization loop searches for the optimal gating coefficients using the loss on the support set, and predictions for the query set are made using the gating coefficients found by the optimization loop. 

For our experiments we retrained SUR's 8 ResNet-18 backbones on their corresponding Meta-Dataset datasets using the original open source codebase and hyper-parameters.
We note that SUR's codebase is affected by a bug that causes the examples of each class to be visited in a deterministic order when sampling episodes. This bug was fixed in our experiments. \footnote{\scriptsize\url{https://github.com/google-research/meta-dataset/issues/54}} This impacts both training and evaluation (Traffic Sign evaluation is particularly sensitive to the issue), which is why our reported baseline accuracies differ from those reported in the original SUR paper (the margin being wider for Traffic Signs).

SUR's preprocessing pipeline resizes the images of Meta-Dataset from their native resolutions to 84 $\times$ 84, using a bilinear interpolation \cite{dvornik2020selecting}. In order to isolate the impact of this processing on aliasing artifacts, we look at its effect on episodic validation performance. As a reminder, validation for each backbone is performed using episodes sampled from its corresponding validation set of classes, which means that the combination of different backbones is eliminated as a potential confounding factor. \autoref{table:sur_val} shows that backbones for which the input data needs heavy downsampling (represented by multiple downward pointing arrows in the {\em Preprocessing} column) benefit the most from the ablation of subsampling in the network's first convolutional layer.

\autoref{table:sur_test} shows the impact of combining anti-aliasing with smooth GELU activation functions on Meta-Dataset's test episode accuracies. Adding low-pass filters on the skip connections yields an average accuracy of 74.80\% (Anti-aliased skip + GELU). Including low-pass filters at all downsampling operations does not significantly improve average performance (74.82\%). We highlight that an average improvement of 3.75\% (absolute) was obtained (including a 2.73\% improvement on out-of-domain tasks). Note that this was achieved with only minor changes to the architecture while using the default hyper-parameters.

 \comment{ 
\section{Stylized-ImageNet: shape \em{versus} texture bias trade-off}

\begin{figure}[ht]
\centering
\begin{minipage}{0.54\linewidth}
{
\includegraphics[width=0.96\linewidth]{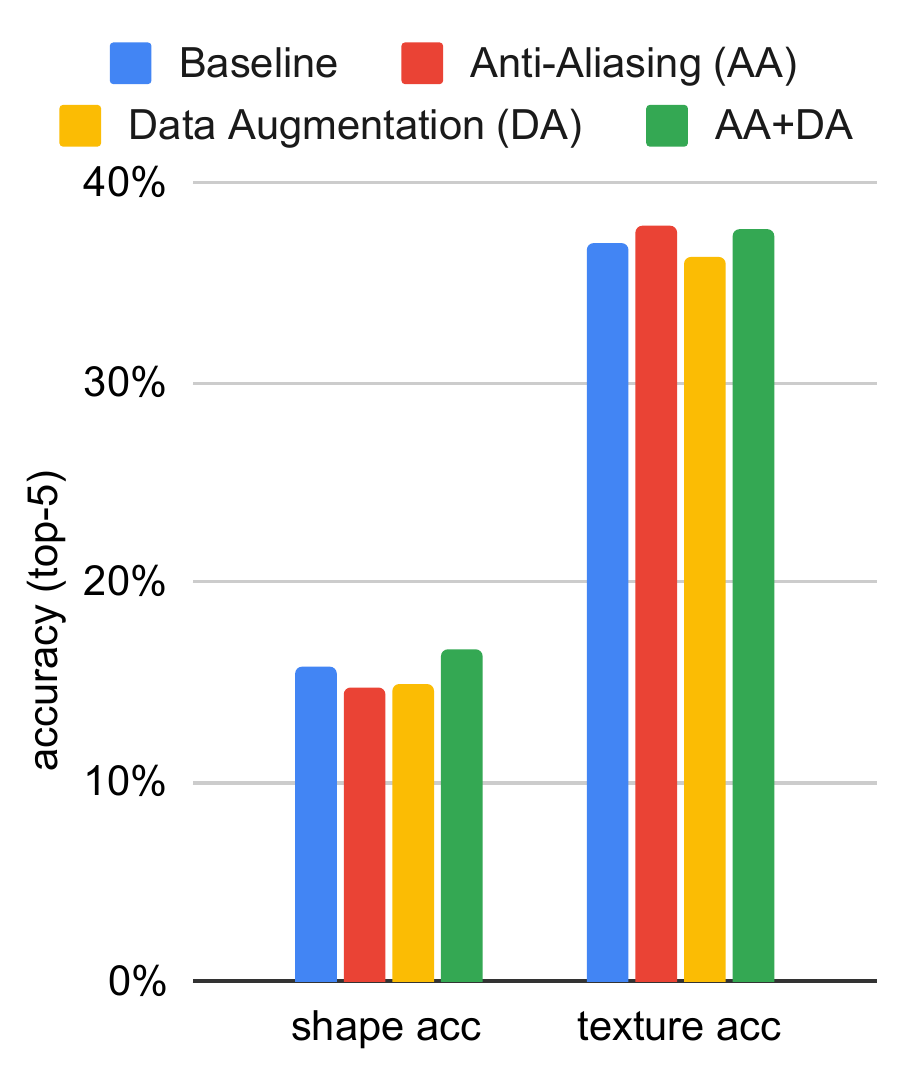}
}
\end{minipage}
\begin{minipage}{0.43\linewidth}
{
\includegraphics[width=0.48\linewidth]{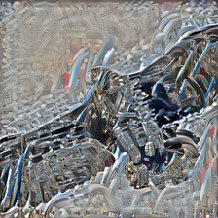}}\hfill
{
\includegraphics[width=0.48\linewidth]{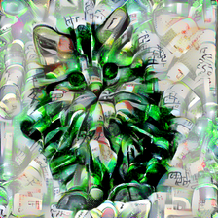}} \\
{
\includegraphics[width=0.48\linewidth]{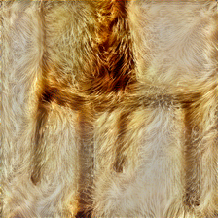}}\hfill
{
\includegraphics[width=0.48\linewidth]{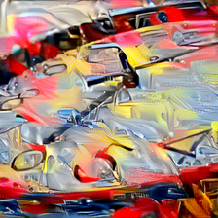}} \\
{
\includegraphics[width=0.48\linewidth]{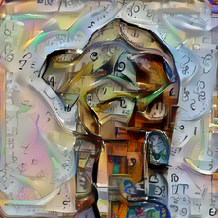}}\hfill
{
\includegraphics[width=0.48\linewidth]{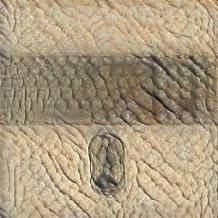}}
\end{minipage}
\caption{Impact on texture-shape cue conflict. {\em Left}:  
Tests on Stylized-ImageNet \cite{geirhos2018imagenettrained}. 
While the anti-aliased model improves texture accuracy, both DA and AA models reduce shape accuracy when taken in isolation, but surpass the baseline when combined.
{\em Right}: sample images containing shapes  correctly classified by the combined model, but by the other three.}
\label{fig:geirhos}
\end{figure}

}

\end{document}